\documentclass[10pt,twocolumn,letterpaper]{article}

\usepackage{cvpr}              % To produce the CAMERA-READY version
\definecolor{cvprblue}{rgb}{0.21,0.49,0.74}
\usepackage[pagebackref,breaklinks,colorlinks,allcolors=cvprblue]{hyperref}
\usepackage{bbm}
\usepackage{dsfont}
\usepackage{amssymb}

\usepackage{multirow}
\usepackage{adjustbox}
\usepackage{colortbl}
\usepackage{booktabs}
\usepackage{algorithm}
\usepackage{algpseudocode}

\def\paperID{*****} % *** Enter the Paper ID here
\def\confName{CVPR}
\def\confYear{2026}

\title{Uncertainty-Aware Exploratory Direct Preference Optimization \\ for Multimodal Large Language Models}

\author{Huatian Zhang \quad Zhendong Mao\thanks{Corresponding author.} \quad Lei Zhang \quad Yongdong Zhang\\
University of Science and Technology of China\\
{\tt\small huatianzhang@mail.ustc.edu.cn \quad \{zdmao,leizh23,zhyd73\}@ustc.edu.cn}
}

\begin{document}
\maketitle
\begin{abstract}

Direct Preference Optimization (DPO) has proven to be an effective solution for mitigating hallucination in Multimodal Large Language Models (MLLMs) by learning from preference pairs. 
One of its key challenges lies in how to transfer the sequence-level preference into fine-grained supervision on visual fidelity. 
To safeguard vision-related tokens that are prone to hallucination, 
existing methods typically allocate training emphasis according to the model's self-assessed visual sensitivity signals.
However, such sensitivity, 
estimated by a model still under training, 
introduces self-referential bias:
reinforcing already well-learned visual cues while neglecting hard-to-perceive but critical details,
thereby limiting deeper alignment.
In this work, we propose an Uncertainty-aware Exploratory Direct Preference Optimization (UE-DPO) method for MLLMs,
which enables the model to uncover its cognitive deficiencies and actively explore for self-correction, guided by token-level epistemic uncertainty.
Specifically, we first quantify the uncertainty from the model's failure to ground token predictions in the given image. 
Then, based on an uncertainty-aware exploration intensity, we encourage more learning pressure on visually deficient tokens in preferred samples,
and alleviate the over-penalization of beneficial knowledge in dispreferred samples. Further, we provide a theoretical justification for our method, and extensive experiments demonstrate its effectiveness and robustness.

\end{abstract}

\section{Introduction}

Multimodal Large Language Models (MLLMs) \cite{dai2023instructblip,liu2024improved,hurst2024gpt,bai2025qwen3,comanici2025gemini} align vision and language modalities by integrating large language model and visual encoder, 
and have demonstrated remarkable visual understanding and reasoning capabilities.
Despite the advancements, 
existing models exhibit a susceptibility to the phenomenon of hallucination,
where the model outputs inaccurately describe or completely fabricate the visual context from input images.
The hallucination problem undermines the reliability of MLLMs and impedes their practical deployment.

\begin{figure}[t]
    \begin{center}
    \includegraphics[width=8.35cm]{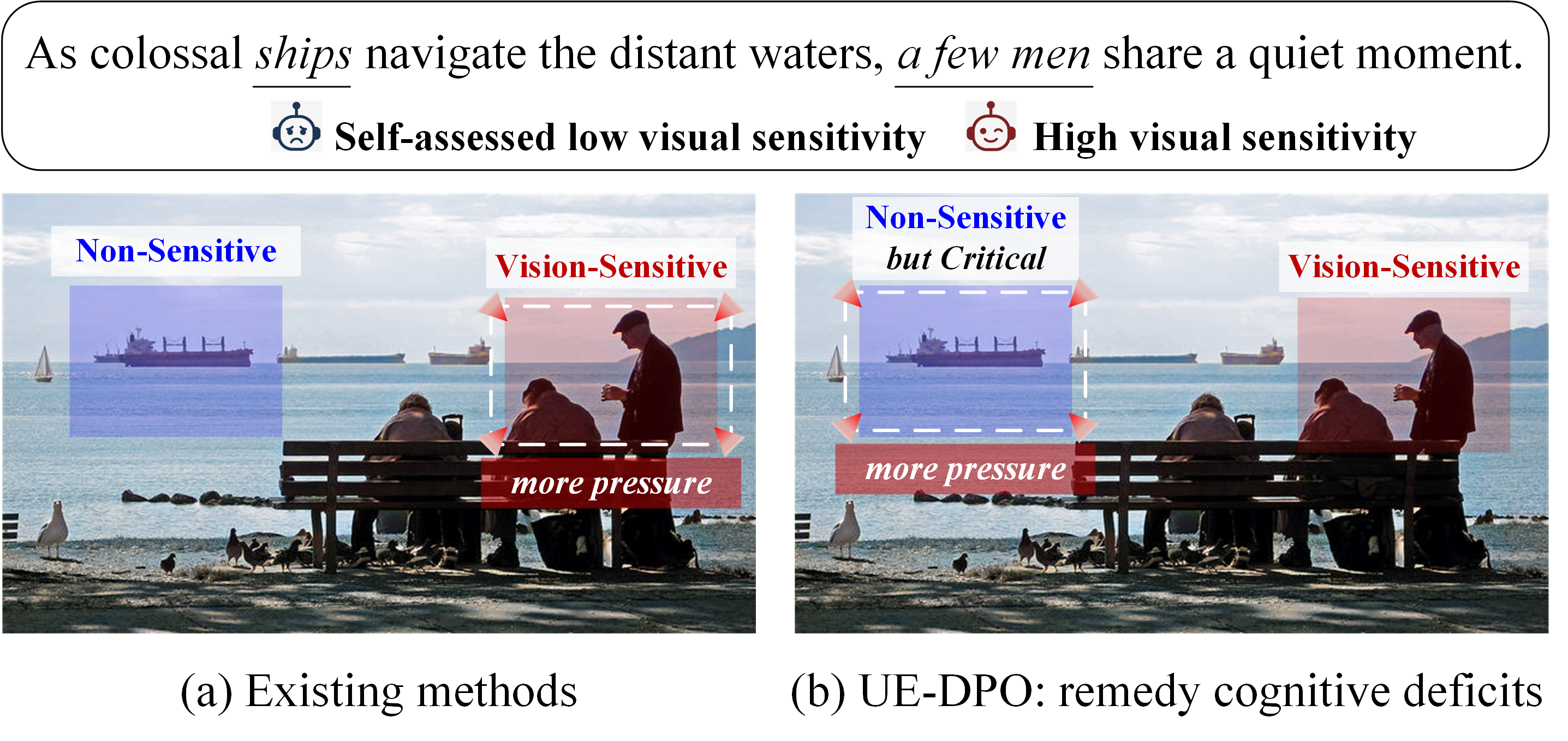}
    \caption{Illustration of the focus shift from established visual sensitivity to cognitive deficits. (a) Existing methods rely on self-assessed visual sensitivity in learning. Stronger optimization pressure is placed on the visually sensitive men, while giving weak pressure to the ``ships'' in response sample, even though such non-sensitive information remains valuable for improving visual understanding. (b) Our UE-DPO method is to rebalance the training by redirecting the focus toward visual cognitive deficiencies. More optimization pressure is applied to insensitive yet crucial ``ships'', to facilitate a deeper alignment.} 
    \label{fig1}	           
    \end{center}
\end{figure}

The hallucinated generation of MLLMs arises from the lack of alignment between vision and language modalities. 
To mitigate this, 
extensive research frames the alignment issue as preference learning and employ Direct Preference Optimization (DPO) \cite{rafailov2023direct} to rectify models toward visual fidelity.
A line of work mitigates hallucinations by constructing effective preference data, such as POVID \cite{zhou2024aligning}, mDPO \cite{wang2024mdpo}, SENTINEL \cite{peng2025mitigating} and OPA-DPO \cite{yang2025mitigating}.
Although the constructed data provides valuable hallucination-contrastive signals,
such sequence-level feedback fails to directly attribute the overall preference to specific tokens,
which limits the learning efficiency.
In light of this,
another line of research focuses on the explicit fine-grained supervision (credit) assignment in preference learning.
For instance,
to give special consideration to vision-related tokens that are susceptible to hallucinations,
TPO \cite{gu2024token} proposed to highlight tokens with a visual-anchored reward in optimization.
V-DPO \cite{xie2024v} integrated a visual guidance factor into learning, derived from the image dependency of token predictions.
In summary,
existing methods typically rely on the self-assessed visual sensitivity signals to improve visual understanding for hallucination mitigation.

However, models that are still under training cannot yet be regarded as fully reliable visual observers, 
as their cognitive capacity remains immature.
The visual information to which the model exhibits sensitivity is, in fact, 
already accessible and can be readily exploited its response generation.
Consequently, further emphasizing such established visual sensitivity to promote visual understanding introduces a potential pitfall:
the model becomes increasingly attuned to familiar visual cues, 
while remaining inattentive to information it cannot effectively interpret,
exposing its cognitive deficiencies.
As a result, 
the key tokens that lack visual sensitivity and embody the model's deficiencies in visual cognition, 
which fundamentally constrain further hallucination mitigation, 
receive insufficient optimization pressure in training.
This observation suggests a critical shift of focus: improving visual understanding should prioritize remedying cognitive deficiencies rather than reinforcing the established visual sensitivity,
as shown in Fig. \ref{fig1}.

To this end, we propose an Uncertainty-aware Exploratory Direct Preference Optimization (UE-DPO) method for MLLMs,
which encourages the model to identify its cognitive deficiencies and engage in self-corrective exploration guided by epistemic uncertainty.
The epistemic uncertainty is quantified by the extent to which the model fails to adequately ground its token predictions in the given visual information.
Specifically, 
we regard the model as exhibiting epistemic uncertainty when, 
despite being provided with clear visual input, 
it assigns lower confidence to vision-related tokens than to those that would be predicted under a stronger reliance on language priors.
In training, greater learning pressure is accordingly imposed on visually under-recognized tokens based on an uncertainty-aware exploration intensity,
while the over-penalization of beneficial visual knowledge is alleviated in the dispreferred samples.
Theoretically, 
our integration of the exploration intensity is equivalent to introducing a token-wise entropy regularization into the reverse-KL regularized RL objective, and implicitly reformulating a generalized advantage function,
which enables the model to break free from the reference policy's visually-deficient prior.
This constitutes a fundamental distinction from the DPO.
Extensive experiments on hallucination benchmarks demonstrate our effectiveness and robustness.

Our contributions are summarized as follows:
\begin{itemize}
 \item We introduce a novel perspective on hallucination mitigation that shifts the emphasis from self-assessed visual sensitivity to addressing cognitive deficiencies, and formalize an epistemic uncertainty measure to identify under-recognized visual information.
 \item We develop an Uncertainty-aware Exploratory Direct Preference Optimization (UE-DPO) method, which adaptively allocates optimization pressure based on an uncertainty-aware exploration intensity to rectify model's visual cognitive deficiencies. A theoretical justification of the proposed objective function is also provided.
 \item Comprehensive evaluations on various hallucination benchmarks demonstrate the effectiveness and robustness of our UE-DPO method.
\end{itemize}

\section{Related Work}

Multimodal large language models (MLLMs) prevalently suffer from serious hallucination problems, in which the generated texts are not factually grounded in associated images.
To alleviate hallucinations, several lines of research focus on generation intervention, 
including post-hoc correction \cite{zhou2023analyzing,yin2024woodpecker,lee2024volcano},
attention rectification \cite{gong2024damro,zhang2024seeing,zhao2025looking,li2025mitigating,zhuang2025vasparse,zhu2025mitigating,che2025hallucinatory,chen2025attention,yin2025clearsight} and training-free refined decoding strategy \cite{leng2024mitigating,zhao2024mitigating,huang2024opera,wang2024mllm,wang2024mitigating,zhu2025ibd,li2025mitigating,li2025hidden,zhang2025self,suo2025octopus}.
Despite these approaches enhance the authenticity of model outputs,
they do not fundamentally address the misalignment between vision and language underlying hallucination.

Preference learning frames the hallucination problem as preference tuning to bias MLLMs towards be non-hallucinated.
A bunch of works \cite{zhao2023beyond,liu2024mia,zhou2024aligning,pi2024strengthening,wang2024mdpo,yu2024rlhf,xie2024v,gu2024token,peng2025mitigating,yang2025mitigating} build upon the Direct Preference Optimization (DPO) \cite{rafailov2023direct}, an stable and effective framework for learning from offline preference data.
For example,
to construct high-quality preference pairs, POVID \cite{zhou2024aligning} and BPO \cite{pi2024strengthening} proposed to directly inject error into ground-truth answers by GPT-4V, as well as create dispreferred samples that expose inherent hallucination bias via blurred images.
mDPO \cite{wang2024mdpo} mitigated the over-prioritization of language through an auxiliary objective conditioned on preference data regarding image completeness.
Inspired by on-policy training,
SENTINEL \cite{peng2025mitigating} bootstraped in-domain preference data by sampling model outputs with progressively enriched contexts.
OPA-DPO \cite{yang2025mitigating} constructed on-policy alignment by fine-tuning model on expert-revised responses,
which makes subsequent preference learning more effective.
Toward explicit credit assignment,
RLHF-V \cite{yu2024rlhf} proposed to learn preference credit allocation based on segment-level labeled correction data.
TPO \cite{gu2024token} and V-DPO \cite{xie2024v} leveraged the self-assessed visual sensitivity signals, 
such as the visual-anchored reward or visual guidance factor in learning for fine-grained hallucination mitigation.

In this work, we aim to shift the focus of hallucination mitigation from reinforcing self-assessed sensitivity toward addressing intrinsic cognitive deficiencies, and to allocate optimization pressure through an uncertainty-aware exploration intensity within the preference learning.

\section{Preliminaries}

\subsection{Token-level MDP for MLLMs}
The response generation process in MLLMs can be formalized as a token-level Markov Decision Process (MDP), denoted as $M=\left(S, A, P, r \right)$.
Within the state space $S$,
each state $s_t$ consists of the input image $v$, the prompt $x$ and the partial output sequence $y_{<t}$ of all tokens generated up to time step $t$,
\emph{i.e.}, $s_t = (v, x, y_{<t} )$, where $y_{<t} = (a_0,\cdots,a_{t-1})$.
The action space $A$ corresponds to the model's vocabulary, 
unlike conventional RL tasks, the action space in large language models is exceptionally large.
The state transition function $P$ is deterministic, that is, 
given the current state $s_t$ and selected token $a_t$, the next state $s_{t+1}$ is obtained by appending $a_t$ to the end of $y_{<t}$.
The reward function $r(s_t, a_t)$ evaluates the quality of taking an token $a_t$ at state $s_t$, 
with rewards that may emerge immediately or be delayed.
It serves as the bridge connecting the MDP to the optimization objectives of RL tasks.

\subsection{The Implicit Advantage View of DPO}
The theoretical foundation of DPO lies in the standard reverse-KL regularized RL framework,
where the learning objective is to maximize:
{\begin{small}
\begin{equation}
J(\pi)=\mathbb{E}_{\tau \sim \pi}\left[\sum_{t=0}^T\left(r\left(s_t, a_t\right)-\beta D_{\mathrm{KL}}\left(\pi\left(\cdot \vert s_t\right) \| \pi_{\mathrm{ref}} \left(\cdot \vert s_t\right)\right)\right)\right].
\label{eq:klrl}
\end{equation}
\end{small}}%
In this setting, the optimal state and action value functions satisfy the Bellman relation:
\begin{equation}
V^*(s)=\max _{\pi(\cdot \mid s)}\left\{\mathbb{E}_{a \sim \pi}\left[Q^*(s, a)\right]-\beta D_{\mathrm{KL}}(\pi \| \pi_{\mathrm{ref}})\right\},
\end{equation}
which couples the trajectory-level objective with single-step decision-making.
The corresponding optimal policy admits a closed-form expression:
\begin{equation}
\pi^*(a \vert s)=\pi_{\mathrm{ref}}(a \vert s) \exp \left(Q^*(s, a) / \beta\right) / Z(s),
\end{equation}
where $Z(s)=e^{V^*(s) / \beta}$, leading to:
\begin{equation}
\beta \log \frac{\pi^*(a \vert s)}{\pi_{\mathrm{ref}}(a \vert s)} = Q^*(s, a) - V^*(s) \triangleq A^*(s, a),
\end{equation}
revealing an explicit correspondence between the policy and its optimal advantage $A^*$,
DPO parameterizes this advantage as an implicit immediate reward \cite{rafailov2024r,knox2024learning,hejnacontrastive} and performs learning by fitting pairwise preferences:
{\begin{small}
\begin{equation}
    \begin{aligned}
    & L_{\text{DPO}}\left(\pi_\theta, \pi_{\mathrm{ref}}\right)=-\mathbb{E}_{\left(x, y_w, y_l\right) \sim \mathcal{D}} \\
    & {\log \sigma\left(\beta \sum_{t=0}^{T_w} \log \frac{\pi_\theta\left(a_t^w \mid s_t\right)}{\pi_{\mathrm{ref}}\left(a_t^w \mid s_t\right)}-\beta  \sum_{t=0}^{T_l} \log \frac{\pi_\theta\left(a_t^l \mid s_t\right)}{\pi_{\mathrm{ref}}\left(a_t^l \mid s_t\right)}\right)},
    \end{aligned}
\label{eq:dpo}
\end{equation}
\end{small}}%
which effectively constrains the learned reward function to belong to the class of optimal advantage function  $A^*$.
This shared structure provides a crucial guarantee:
in learning, as the parameterized $A_{\theta}$ is optimized to approximate $A^*$, 
the policy distribution is correspondingly adjusted, ensuring that the underlying policy $\pi_\theta$ inevitably approaches $\pi^*$. 
The refinement of the value landscape of the parameterized advantage $A_{\theta}$ through Eq. \ref{eq:dpo} is thus faithfully mirrored in the shaping of the policy $\pi_\theta$.
DPO is able to flexibly model any possible dense reward function \cite{rafailov2024r} within the token-level MDP.
See Appendix A for details.

\section{Methodology}

We then elaborate on the framework of UE-DPO, which aims to remedy cognitive deficiencies of MLLMs by explicitly modeling token-level epistemic uncertainty and adaptively regulating exploration intensity on response samples in preference learning. A schematic overview is shown in Fig. \ref{fig2}, and algorithmic outline is provided in Appendix B.

\subsection{Uncertainty Awareness}
\label{uncertainty}
Epistemic uncertainty \cite{kuhn2023semantic, yadkori2024believe, ma2025estimating} arises from the lack of knowledge about the ground truth.
To identify under-recognized visual information,
we measure the epistemic uncertainty of model in response generation,
which reflects the extent to which token predictions fail to be adequately grounded in visual evidence.
The policy $\pi_\theta$ is deemed to exhibit epistemic uncertainty if,
given a clear image $v$,
it assigns lower confidence (logit) to the ground-truth vision-related token $a_t$ than to the token $\hat{a}_t(v^\prime)$ predicted from a blurred image $v^\prime$, where language priors are more likely to dominate.
Specifically,
we measure epistemic uncertainty at the time step $t$ as:
\begin{equation}
    \mathrm{u}(s_t, a_t) = \mathrm{logit}_\theta(\hat{a}_t(v^\prime) \vert v,x,y_{<t}) - \mathrm{logit}_\theta(a_t \vert v,x,y_{<t}),
\label{eq:unc}
\end{equation}
where $\hat{a}_t(v^\prime) = \arg \max \pi_\theta(\cdot \vert v^\prime,x,y_{<t})$, and the blurred image $v^\prime$ is obtained by adding diffusion noise \cite{ho2020denoising},
which can be represented as:
\begin{equation}
v^{\prime}(k)=\sqrt{\bar{\xi}_k} \cdot v + \sqrt{1-\bar{\xi}_k} \cdot \epsilon,
\end{equation}
where $\bar{\xi}_k=\prod_{i=0}^k \xi_i$ and $\xi \in (0,1)$ is a hyper-parameter governing the noise level,
and $\epsilon$ is randomly sampled from a normal distribution for each sample.
A high level of epistemic uncertainty indicates that the image does not provide sufficient positive and vision-faithful support for token prediction in response generation,
reflecting the model's limited grounding in the relevant visual content.

\subsection{The Control of Exploration Intensity}
\label{sec:Exploration}
We adopt an asymmetric strategy to allocate token-level visual exploration intensity across preferred and dispreferred samples, guided by the estimated uncertainty.

\noindent \textbf{Preferred Branch.} For visually insensitive tokens in the preferred samples $y_w=(a_0,\dots,a_{T})$ that predominantly rely on language priors during prediction,
\emph{i.e.}, tokens whose logit variations, $\Delta(s_t, a_t) = \mathrm{logit}_\theta(a_t \vert v,x,y_{<t}) - \mathrm{logit}_\theta(a_t \vert v^\prime,x,y_{<t})$, are minimal or even positive after the image is blurred,
high epistemic uncertainty indicates that the model ideally ought to have leveraged visual information in prediction, yet instead defaulted to language priors due to insufficient grounding in visual content.
Conversely, when a visually insensitive token exhibits low uncertainty, its reliance on linguistic priors can be regarded as contextually appropriate and semantically justified.
Specifically, we indicate visually insensitive tokens by:
\begin{equation}
\mathrm{I}_w=\mathbbm{1}\left\{\Delta(s_t, a_t) \leq \mathrm{q}_{\tau}(\Delta(s_t, a_t))\right\}, t = 0,\dots,T,
\end{equation}
where $\mathbbm{1}\{\cdot\}$ denotes indicator function,
and $\mathrm{q}_{\tau}(\cdot)$ denotes the $\tau$ quantile.
Then we determine the exploration intensity for tokens within preferred samples as:
\begin{equation}
\lambda_w(s_t, a_t)=1 + \alpha \mathbbm{1}\left\{ \mathrm{I}_w = 1\right\} \sigma(\frac{\mathrm{u}(s_t, a_t) - \mu_I }{\varsigma_I}),
\label{eq:prefcont}
\end{equation}
where $\sigma(\cdot)$ denotes the sigmoid function, $\alpha$ is a hyper-parameter controlling the intensity scale, $\mu_I$ and $\varsigma_I$ represent the first quantile and standard deviation of the uncertainty associated with tokens satisfying $\mathrm{I}_w = 1$, respectively.
Eq. \ref{eq:prefcont} enables a targeted modulation of the exploration intensity on the preferred samples:
it adaptively encourages exploration for high-uncertainty ``guessed'' tokens where the visual understanding is insufficient,
while maintaining stability for the low-uncertainty tokens that are confidently predicted,
thereby avoiding interference with the model's legitimate use of language priors.

\noindent \textbf{Dispreferred Branch.}
Not all elements within dispreferred responses are inconsistent with the visual input.
For visually sensitive tokens in the dispreferred samples $y_l=(a_0,\dots,a_T)$,
relatively high epistemic uncertainty indicates that policy $\pi_\theta$ tends to prioritize tokens predicted predominantly by language priors over vision-related ones.
In preference learning,
imposing penalties on dispreferred samples may further attenuate the model's visual cognition.
Hence, we selectively mitigate the penalty based on the estimated uncertainty in training.
Specifically, visually sensitive tokens are identified as
\begin{equation}
\mathrm{I}_l=\mathbbm{1}\left\{\Delta(s_t, a_t) \geq \mathrm{q}_{1-\tau}(\Delta(s_t, a_t))\right\}, t = 0,\dots,T,
\end{equation}
where $\mathrm{q}_{1-\tau}(\cdot)$ denotes the $1-\tau$ quantile.
Then, we set the token-wise mitigation on the penalty as:
\begin{equation}
\lambda_l(s_t, a_t)=1 - \alpha \mathbbm{1}\left\{ \mathrm{I}_l = 1\right\} \sigma(\frac{\mathrm{u}(s_t, a_t) - \mu_I }{\varsigma_I}),
\label{eq:disprefcont}
\end{equation}
where $\mu_I$ and $\varsigma_I$ indicate the first quantile and the standard deviation of the uncertainties with $\mathrm{I}_l = 1$, respectively.

Eq. \ref{eq:disprefcont} adaptively scales down the penalty in proportion to the relative magnitude of epistemic uncertainty for visually sensitive tokens, \emph{i.e.}, the relative risk of cognitive degradation.
This mechanism prevents preference learning from forgetting the established visual cognitions,
while simultaneously acting as a safeguard for policy exploration: by conservatively allocating penalty,
we leave room for the model to continue exploring these pieces of knowledge.

\begin{figure}[t]
    \begin{center}
    \includegraphics[width=7.8cm]{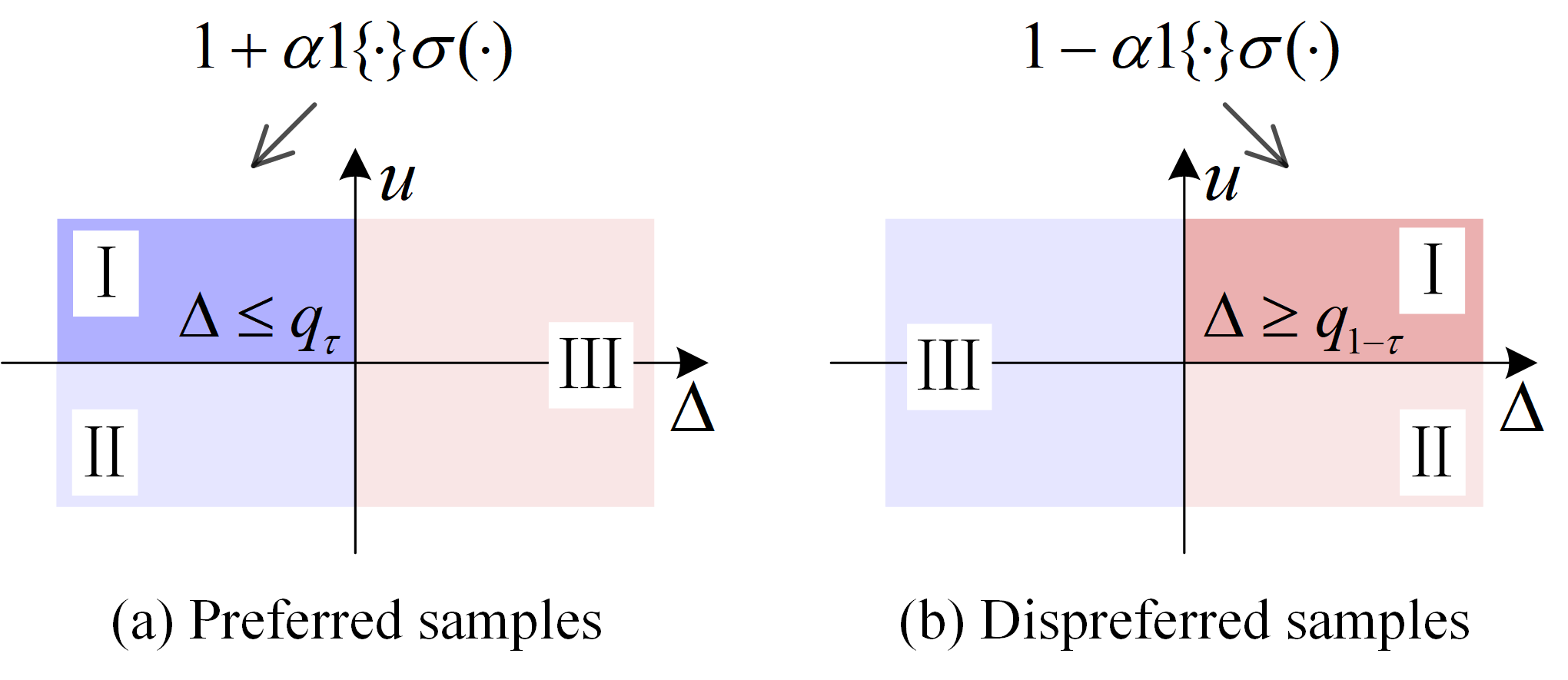}
    \caption{Schematic illustration of our method. (a) For preferred responses, the method intensifies exploration on Type-\uppercase\expandafter{\romannumeral1} tokens characterized by high uncertainty and low sensitivity.
    Type-\uppercase\expandafter{\romannumeral2} tokens correspond to legitimate language dependencies.
    Type-\uppercase\expandafter{\romannumeral3} tokens already exhibit high visual sensitivity. 
    (b) For dispreferred responses, the method mitigates preference penalties on Type-\uppercase\expandafter{\romannumeral1} tokens exhibiting both high epistemic uncertainty and visual sensitivity. The visual grounding of Type-\uppercase\expandafter{\romannumeral2} tokens is sufficiently stable, and Type-\uppercase\expandafter{\romannumeral3} tokens are visually insensitive in our view.} 
    \label{fig2}	           
    \end{center}
\end{figure}

\subsection{Training Objective}
\label{sec:training}
We integrate the token-level exploration controlling into the DPO framework as:
{\begin{small}
\begin{equation}
    \begin{aligned}
    & L_{\text{UE-DPO}}\left(\pi_\theta, \pi_{\mathrm{ref}}\right)=-\mathbb{E}_{\left(x, y_w, y_l\right) \sim \mathcal{D}} \\
    & {\log \sigma\left(\beta \sum_{t=0}^{T_w} \log \frac{\pi_\theta\left(a_t^w \vert s_t\right)^{\operatorname{sg}[\lambda_w]}}{\pi_{\mathrm{ref}}\left(a_t^w \vert s_t\right)} -\beta  \sum_{t=0}^{T_l} \log \frac{\pi_\theta\left(a_t^l \vert s_t\right)^{\operatorname{sg}[\lambda_l]}}{\pi_{\mathrm{ref}}\left(a_t^l \vert s_t\right)}\right)},
    \end{aligned}
\label{eq:uedpo}
\end{equation}
\end{small}}%
which adaptively regulates the implicit advantages in Eq. \ref{eq:dpo} through our controlled exploration intensity built upon epistemic uncertainty.
The $\operatorname{sg}[\cdot]$ denotes the operation of stopping gradient  on $\lambda_w(s_t, a_t)$ and $\lambda_l(s_t, a_t)$.
In training,
this objective drives the model to focus more on visually under-cognized tokens in preferred samples through the rescaled gradient $\lambda_w(s_t, a_t) \nabla_\theta \log \pi_\theta\left(a_t \vert s_t\right)$,
and safeguards the model's acquired knowledge in dispreferred samples,
through the gradient $\lambda_l(s_t, a_t) \nabla_\theta \log \pi_\theta\left(a_t \vert s_t\right)$.
It achieves a principled balance between promoting uncertainty-driven visual exploration and maintaining the integrity of pre-trained knowledge via fine-grained, uncertainty-aware credit assignment in preference learning.

\section{Theoretical Justification}

The integration of the uncertainty-aware exploration can be interpreted as introducing a token-wise entropy regularization factor $\lambda(s,a)$ into the reverse-KL regularized RL objective in Eq. \ref{eq:klrl}.
This factor dynamically modulates the KL regularization strength in accordance with the model's epistemic uncertainty across different tokens, leading to the following reformulated value function:
\begin{equation}
\max _{\pi(\cdot \vert s)}\{\mathbb{E}_{a \sim \pi}\left[Q^*(s,a) - \beta (\lambda \log \pi(a \vert s) - \log \pi_{\mathrm{ref}}(a \vert s))\right]\}.
\label{eq:ueobj}
\end{equation}
In the following, we analyze how the exploration intensity factor $\lambda$ affects the model from two aspects.

\noindent \textbf{The influence of $\lambda$ on the optimal policy.}
The optimal policy of Eq. \ref{eq:ueobj} can be derived as:
\begin{equation}
\pi^{*}(a \vert s)=\pi_{\mathrm{ref}}(a \vert s)^{1 / \lambda} \exp \left( \frac{Q^{*}(s,a) + \eta(s)}{\beta \lambda}\right) / Z(s),
\label{eq:revisedobj}
\end{equation}
where $Z(s)$ is partition function, and $\eta$ is Lagrange multiplier.
The derivative of $\log \pi^{*}$ with respect to $\lambda$ is proportional to:
$- \log(\pi_{\mathrm{ref}}(a|s)) / \lambda^2 - (Q^{*}(s,a) + \eta(s)) / (\beta \lambda^2)$,
% \begin{equation}
% \frac{\partial}{\partial \lambda} \log \pi^{*} \propto  -\frac{\log \left(\pi_{\mathrm{ref}}(a \vert s)\right)} {\lambda^2} - \frac{\left(Q^*(a, s)+\eta(s)\right)}{\left(\beta \lambda^2\right)},
% \end{equation}
where the first term is always positive, and its magnitude increases as $\pi_{\mathrm{ref}}(a \vert s)$ decreases.
Of particular interest is the case relevant to hallucination mitigation, namely vision-related tokens subject to under-cognition, \emph{i.e.}, those high-value tokens characterized by large $Q^*(s,a)$ and low reference probabilities $\pi_{\mathrm{ref}}$.
In such cases, a larger $\lambda$ exerts a pronounced corrective influence on $\pi_{\mathrm{ref}}(a \vert s)$ within the optimized policy $\pi^*$, effectively counteracting the smoothing effect introduced by $\lambda$ via $\exp (\cdot)$ factor.
Consequently, the target policy can deviate from suboptimal constraints imposed by the reference model, 
thereby achieving a directed correction of cognitive bias.
See Appendix C for details.

\noindent \textbf{Generalized Exploratory Advantage.}
It follows that the optimal policy $\pi^{*}(a \vert s)$ ( in Eq. \ref{eq:revisedobj}) satisfies:
\begin{equation}
\begin{aligned}
&\beta \log \frac{\pi^*(a \vert s)^{\lambda}}{\pi_{\mathrm{ref}}(a \vert s)} \\
& = Q^{*}(s,a)-V^{*}(s)-\beta (\lambda -\mathbb{E}_{a^{\prime} \sim \pi^*}\left[\lambda^\prime\right]) \\
& \triangleq A_\mathrm{e}^*(s,a),
\end{aligned}
\label{eq:condition}
\end{equation}
where $V^{*}(s) =\beta \mathbb{E}_{a \sim \pi^*}[\lambda]-\eta(s) $, which is the solution of the objective in Eq. \ref{eq:ueobj}.
The presence of $\lambda$ leads to an extended form of the advantage function,
which we refer to as the generalized exploratory advantage $A_\mathrm{e}^*$:
as revealed by Eq. \ref{eq:condition},
this advantage function can be decomposed into the conventional value advantage $Q^{*}(s,a)-V^{*}(s)$ and exploration cost $\beta (\lambda -\mathbb{E}_{a^{\prime} \sim \pi^*}\left[\lambda^\prime\right])$.
That is, this advantage is a static evaluation of the policy that also includes exploration intensity as a cost.
%This advantage is a static evaluation of the policy with uncertainty-aware cognition risk awareness.
As $\lambda$ diminishes in training, the advantage increases.
By substituting this generalized exploratory advantage into the DPO framework,
our proposed UE-DPO can fit the pair-wise preference ordering while effectively mitigating the under-cognition of visual information through the gradients weighted by $\lambda$ for exploration (Sec. \ref{sec:training}).
See Appendix D for details.

\section{Experiments}

\subsection{Setup}

\subsubsection{Models and Datasets}
We conduct experiments on various models LLAVA-v1.5-7B \cite{liu2024improved}, LLAVA-v1.5-13B \cite{liu2024improved} and Qwen2.5-VL-3B \cite{bai2025qwen2}.
LLAVA-v1.5-7B and LLAVA-v1.5-13B are built upon Vicuna-7B and Vicuna-13B, respectively, and utilize CLIP ViT-L-336px as visual encoder.
Qwen2.5-VL are based on the Qwen2.5 LLM and a redesigned ViT.
We conduct experiments on the public human-feedback preference dataset RLHF-V \cite{yu2024rlhf} and AI-feedback dataset RLAIF-V \cite{yu2025rlaif}.

\subsubsection{Benchmarks}
We evaluate our model on various hallucination benchmarks.
MMHal-Bench \cite{sun2024aligning} is a multimodal question-answering benchmark designed to evaluate the informativeness and hallucination rate of model responses. 
It includes 96 images covering 12 object categories and 8 question types. 
Following the official protocol, GPT-4 is used to rate model outputs on a 0-6 scale, where responses scoring below 3 are considered hallucinated.
Object HalBench \cite{rohrbach2018object} is a benchmark widely used to evaluate object hallucination in vision-language models.
We follow the setting of \cite{yu2025rlaif} to evaluate on 300 instances with diverse prompts.
It quantifies the hallucination rate at both the response level CHAIRs and the granular object level CHAIRi.
AMBER \cite{wang2023llm} contains both generative and discriminative tasks:
AMBER-g focuses on evaluating models' ability to generate faithful and descriptive captions for 1,004 images with detailed object-level annotations. We report CHAIR score, hallucination rate, and the alignment with human cognition (Cog.).
AMBER-d targets to identify hallucinated content through yes/no question-answering. It contains over 14k samples and evaluates performance using Accuracy and F1 metrics.

\subsubsection{Implementation Details}
In training, we set the maximum learning rate as 1e-5 with a cosine annealing scheduler, and the batch size is $64$ with gradient accumulation.
The models are trained for $2$ epochs on both RLHF-V and RLAIF-V dataset.
We adopt LoRA \cite{hu2022lora} for training with the LoRA rank as $128$.
The hyper-parameter $\alpha$ controlling the exploration intensity is set to $0.3$ for LLAVA-v1.5-7B, $0.25$ for  LLAVA-v1.5-13B, and $0.15$ for Qwen2.5-VL-3B.
Following previous works, the hyper-parameter $\beta$ in the reparameterized reward function is set to $0.1$.
The diffusion noise level $\xi = \sigma \left(l_t\right) \cdot \left(0.5 \times 10^{-2}-10^{-5}\right)+10^{-5}$,
where $l_t$ is a list of $1,000$ numbers taken at equal intervals over the interval $[-6, 6]$, and the noise step $k$ is set to $500$.
Our experiments are conducted on up to 4 NVIDIA A100 GPUs.
The implementation code is publicly available \footnote{https://github.com/htzhang-code/UE-DPO}.

\begin{table*}[ht]
\centering
\caption{Performance comparison with preference learning based hallucination mitigation methods for MLLMs across various hallucination benchmarks. For baselines without official checkpoints, the results are taken from the respective papers. For baselines with available official checkpoints, we primarily refer to the re-evaluation results reported in \cite{yang2025mitigating}. The $\dagger$ indicates that the method was trained on the same dataset as ours. The best for each metric is in bold.}
\renewcommand{\arraystretch}{1.05}
\setlength{\tabcolsep}{4pt}
\small
\begin{adjustbox}{max width=\textwidth}
%\resizebox{\textwidth}{!}{
\begin{tabular}{lcccccccccc}
\toprule
\multirow{2}{*}{\textbf{Method}} & \multicolumn{1}{c|}{\multirow{2}{*}{\textbf{Datasize}}} & \multicolumn{2}{c}{\textbf{Object-Hal}}           & \multicolumn{2}{c}{\textbf{MMHal-Bench}}                & \multicolumn{3}{c}{\textbf{AMBER-g}}                                        & \multicolumn{2}{c}{\textbf{AMBER-d}}                 \\ \cmidrule(r){3-4} \cmidrule(lr){5-6} \cmidrule(lr){7-9} \cmidrule(lr){10-11}  
                        & \multicolumn{1}{c|}{}        & CHAIRs$\downarrow$         & CHAIRi$\downarrow$   & Score$\uparrow$         & HalRate$\downarrow$         & CHAIR$\downarrow$   & HalRate$\downarrow$   & Cog.$\downarrow$   & Acc.$\uparrow$          & F1$\uparrow$          \\ \midrule
\textbf{LLaVA-v1.5-7B}            &  \multicolumn{1}{c|}{} & 55.67              & 15.96   & 2.01             & 0.61             & 7.7       & 34.7         & 4.2     & 71.7             & 74.3           \\
+ LLaVA-RLHF \cite{sun2024aligning}          &  \multicolumn{1}{c|}{122k}   & 58.00              & 15.61             & 1.88             & 0.71              & 9.7       & 46.6         & 5.3     & --            & --           \\
+ HALVA \cite{sarkar2024mitigating}           &  \multicolumn{1}{c|}{21.5k}  & 41.40              & 11.70             & 2.25             & 0.54               & 6.6       & 32.2         & 3.4     & --             & --           \\ 
+ POVID \cite{zhou2024aligning}           &  \multicolumn{1}{c|}{17k}  & 50.67              & 15.28             & 2.08             & 0.60               & 7.4       & 34.3         & 3.9     & 71.9             & 74.7           \\
+ mDPO  \cite{wang2024mdpo}          &  \multicolumn{1}{c|}{10k}  & 35.70              & 9.80             & 2.39             & 0.54               & 4.4       & 24.5         & 2.4     & --             & --           \\
+ V-DPO$^\dagger$ \cite{xie2024v}           & \multicolumn{1}{c|}{5.7k}  & --              & --             & 2.16             & 0.56               & 5.6       & 27.3         & 2.7     & --             & 81.6           \\
+ TPO$^\dagger$ \cite{gu2024token}          & \multicolumn{1}{c|}{5.7k}   & --              & --             & 2.47             & 0.51               & --       & --         & --     & 79.3             & 85.0           \\
+ RLAIF-V  \cite{yu2025rlaif}           & \multicolumn{1}{c|}{16k}   & 16.0              & \textbf{3.70}             & \textbf{3.00}             & 0.38               & 3.0       & 16.2         & 1.0    & --             & --           \\
\rowcolor{gray!20}
\emph{on RLHF-V Data}          & \multicolumn{1}{l}{}                           & \multicolumn{1}{l}{} & \multicolumn{1}{l}{} & \multicolumn{1}{l}{} & \multicolumn{1}{l}{} & \multicolumn{1}{l}{} & \multicolumn{1}{l}{} & \multicolumn{1}{l}{} & \multicolumn{1}{l}{} & \multicolumn{1}{l}{} \\
\rowcolor{gray!20}
+ \textbf{UE-DPO$^\dagger$ (ours)}           & \multicolumn{1}{c|}{5.7k}                         & 13.72                    & 6.69                    & 2.82                    & 0.48                    & 2.9                    & 17.4                    & 1.0                    & 79.1                    & 85.7                    \\ 
\rowcolor{gray!20}
\emph{on RLAIF-V Data}          & \multicolumn{1}{l}{}                           & \multicolumn{1}{l}{} & \multicolumn{1}{l}{} & \multicolumn{1}{l}{} & \multicolumn{1}{l}{} & \multicolumn{1}{l}{} & \multicolumn{1}{l}{} & \multicolumn{1}{l}{} & \multicolumn{1}{l}{} & \multicolumn{1}{l}{} \\
\rowcolor{gray!20}
+ \textbf{UE-DPO (ours)}            & \multicolumn{1}{c|}{16k}                         & \textbf{11.62}                    & 5.16                    & 2.95                    & \textbf{0.37}                    &  \textbf{2.5}                    & \textbf{12.5}                    & \textbf{0.9}                    & \textbf{81.7}                    & \textbf{87.0}                   \\\midrule
\textbf{LLaVA-v1.5-13B}            & \multicolumn{1}{c|}{} & 51.00          & 13.71          & 2.48                    & 0.52                    & 6.8                    & 31.8                    & 3.3                    & 71.3                    & 73.1                    \\
+ LLaVA-RLHF \cite{sun2024aligning}          & \multicolumn{1}{c|}{122k}   & 44.67  & 11.83                     & 2.27                    & 0.64                    & 7.7                    & 38.6                    & 4.0                    & --                    & --                    \\
+ HALVA \cite{sarkar2024mitigating}            & \multicolumn{1}{c|}{22.5k}    & 45.40                    & 12.80       & 2.58                    & 0.45                    & 6.4                    & 30.4                    & 3.2                    & --                    & --                    \\
+ RLHF-V(HD) \cite{yu2024rlhf}          & \multicolumn{1}{c|}{1.4k}   & --  & --                     & \textbf{2.81}                    & 0.49                    & 6.3                    & 25.1                    & 2.1                    & --                    & --                    \\
+ TPO$^\dagger$  \cite{gu2024token}     & \multicolumn{1}{c|}{5.7k}  & --              & --    & 2.72             & 0.46        & --       & --         & --     & \textbf{83.9}             & 88.0           \\
\rowcolor{gray!20}
+ \textbf{UE-DPO$^\dagger$ (ours)}           & \multicolumn{1}{c|}{5.7k}                         & \textbf{13.70}                    & \textbf{6.57}                    & 2.76                    & \textbf{0.45}                    & \textbf{2.2}                    & \textbf{14.4}                    & \textbf{0.9}                    & 83.5                    & \textbf{88.2}                    \\ \midrule
\textbf{Qwen2.5-VL-3B}            & \multicolumn{1}{c|}{} &  43.2          & 12.2          & 2.62                    & 0.49                    & 8.0                    & 44.4                    & 4.3                    &  \textbf{90.0}                    & 86.0                    \\
+ DPO$^\dagger$         & \multicolumn{1}{c|}{5.7k}   & 30.6  & 10.7                     & 2.53                    & 0.52                    & 5.7                    & 22.6                    & 1.4                    & 86.2                    & 88.9                    \\
\rowcolor{gray!20}
+ \textbf{UE-DPO$^\dagger$ (ours)}           & \multicolumn{1}{c|}{5.7k}    & \textbf{16.7}                    &  \textbf{5.6}       &  \textbf{3.05}                    &  \textbf{0.41}                    &  \textbf{3.0}                    &  \textbf{14.7}                    &  \textbf{0.6}                    & 85.1                    &  \textbf{89.5}                    \\ \bottomrule
\end{tabular}
%}
\label{tab:1}
\end{adjustbox}
\end{table*}

\subsubsection{Baselines}

We primarily compare UE-DPO with existing hallucination mitigation methods based on preference learning,
including POVID \cite{zhou2024aligning}, mDPO \cite{wang2024mdpo}, HALVA \cite{sarkar2024mitigating}, RLHF-V \cite{yu2024rlhf}, RLAIF-V \cite{yu2025rlaif},
particularly the credit self-estimation method V-DPO \cite{xie2024v} and TPO \cite{gu2024token}.
To ensure a fair comparison, 
methods that rely on intricate preference data construction procedures,
such as SENTINEL \cite{peng2025mitigating} and OPA-DPO \cite{yang2025mitigating},
are excluded from direct comparison.
Notably, UE-DPO demonstrates competitive performance relative to these approaches across most evaluation metrics.

\subsection{Main Results}

Our method exhibits clear performance superiority across most evaluation metrics and model scales, as shown in Tab. \ref{tab:1}.
With applying to multiple MLLM backbones,
the consistently substantial improvements highlights the effectiveness of our uncertainty-aware exploratory method in reducing hallucinations compared to baselines involving various preference learning strategies.

In terms of Object-Hal, our method achieves the lowest CHAIRs and CHAIRi scores on all tested backbones, confirming its superior ability to suppress object-level hallucinations. 
On LLaVA-v1.5-7B, we reduce the scores to a notably lower level under both RLHF-V and the larger-scale RLAIF-V datasets,
while the 13B model also maintains a low hallucination rate.
On Qwen2.5-VL-3B, the CHAIRs score exhibits a substantial reduction.
The consistent improvements substantiate that our method effectively enforces more precise object grounding and constrains undesired response-generations at the object recognition level.

For the MMHal-Bench,
our method achieves the highest score and the lowest HalRate across nearly all baselines.
As a case in point, 
on the LLaVA-v1.5-7B model trained with the RLHF-V dataset,
we significantly outperform the similar credit self-estimation methods such as TPO and V-DPO trained on the same data.
These demonstrate that, in the MMHal-Bench evaluation, our method achieves a more favorable balance between mitigating hallucinations and maintaining the accuracy of generated content.

On the AMBER-g,
the results suggest that our method demonstrates strong capability in handling diverse and fine-grained hallucination scenarios.
Regarding AMBER-d,
our method maintains a leading position on F1, but exhibits slightly lower performance on Acc. metric.
In particular,
when trained on the RLHF-V dataset, the Acc. slightly decreases.
A key goal of hallucination mitigation is to prevent fabrication. AMBER-d is a discriminative (yes-or-no) benchmark, where Acc. is calculated on all questions, and F1 is only on hallucinatory questions (ground truth is ``no'').
Our higher F1 implies the improved mitigation (suppressing fabrication), and the lower Acc. reflects a trade-off that the model has become cautious for credibility, providing fewer ``yes'' answers, thus reducing true positives then lowering accuracy.
While using the larger and broader RLAIF-V dataset leads to a clear rebound that both F1 and Acc. increases.
This suggests that AMBER-d may be highly sensitive to data scale and coverage.
% Therefore, expanding the preference data with broader and more diverse coverage is likely to further improve Acc without compromising F1. 
Overall, the results on AMBER-d indicate that while our framework remains competitive, 
there is still room for improvement: 
further enhancement of dataset coverage could yield stronger performance in detail-sensitive hallucination detection.

\begin{table}[t]
\centering
\caption{Ablation study of exploration control on the preferred and dispreferred branches with LLaVA-v1.5-7B. The w/o pref. indicates that exploration control is retained only on the dispref. branch, while the pref. branch reverts to the standard DPO log-probability summation. Conversely, the w/o dispref. applies control only to the pref. branch. The complete UE-DPO method applies exploration control to both branches.}
\renewcommand{\arraystretch}{1.02}
\setlength{\tabcolsep}{1.pt}
\small
%\begin{adjustbox}{max width=\textwidth}
%\resizebox{\textwidth}{!}{
\begin{tabular}{lccccc}
\toprule
\multirow{2}{*}{\textbf{Method}} & \multicolumn{2}{c}{\textbf{MMHal-Bench}} & \multicolumn{3}{c}{\textbf{AMBER}} \\ \cmidrule(r){2-3} \cmidrule(lr){4-6} 
                        & Score$\uparrow$         & HalRate$\downarrow$        & CHAIR$\downarrow$   & HalRate$\downarrow$   & Cog$\downarrow$   \\ \midrule
\emph{on RLHF-V Data}            &                &               &         &           &       \\
DPO                     & 2.26              & 0.60             & 3.7       & 22.9         & 1.6     \\
w/o pref.           & 2.51              & 0.55             & 3.6       & 21.7         & 1.5     \\
w/o dispref.        & 2.73              & 0.50             & 2.8       & 17.1         & 1.0     \\
\rowcolor{gray!20}
UE-DPO                  & 2.82              & 0.48             & 2.9       & 17.4         & 1.0     \\ \midrule
\emph{on RLAIF-V Data}           &                &               &         &           &       \\
DPO                     & 2.73              & 0.38             & 2.7       & 13.4         & 1.1     \\
w/o pref.           & 2.65              & 0.40             & 2.7       & 14.1         & 1.2     \\
w/o dispref.        & 2.84              & 0.37             & 2.6       & 12.8         & 1.0     \\
\rowcolor{gray!20}
UE-DPO                  & 2.95              & 0.37             & 2.5       & 12.5         & 0.9     \\ \bottomrule
\end{tabular}
\label{tab:2}
%}
%\end{adjustbox}
\end{table}

\subsection{Ablation Studies}

\noindent \textbf{Exploration on Preferred and Dispreferred Branches.}
Our exploration control consists of two complementary branches: 
an exploratory incentive on preferred samples, 
encouraging focus on visual cues that exhibit high epistemic uncertainty yet low sensitivity,
and a protective penalty alleviation on dispreferred samples, 
which mitigates over-penalization of highly sensitive visual knowledge.
The ablation study on the preferred and dispreferred branches examines their respective contributions to hallucination mitigation, as shown in Tab. \ref{tab:2}. 
The results indicate that the preferred branch serves as the primary driver for hallucination reduction:
using this branch alone (w/o dispref.) yields a substantial performance leap over the baseline DPO.
Meanwhile, 
the penalty alleviation on the dispreferred branch provides an auxiliary effect.
Its effect alone remains relatively weak.
When both branches are jointly applied (UE-DPO), the model achieves further improvements across most metrics.
These findings confirm that integrating exploration control over both branches leads to a more balanced and robust  mitigation of hallucinations.

\begin{table}[t]
\centering
\caption{Ablation study on the exploration intensity factor $\alpha$.}
\renewcommand{\arraystretch}{1.02}
\setlength{\tabcolsep}{1pt}
\small
%\begin{adjustbox}{max width=\textwidth}
%\resizebox{\textwidth}{!}{
\begin{tabular}{lccccc}
\toprule
\multirow{2}{*}{\textbf{Method}} & \multicolumn{2}{c}{\textbf{MMHal-Bench}} & \multicolumn{3}{c}{\textbf{AMBER}} \\ \cmidrule(r){2-3} \cmidrule(lr){4-6} 
                        & Score$\uparrow$         & HalRate$\downarrow$        & CHAIR$\downarrow$   & HalRate$\downarrow$   & Cog$\downarrow$   \\ \midrule
\emph{Exploration factor}            &                &               &         &           &       \\
$\alpha=0.20$                     & 2.70              & 0.51             & 3.2       & 18.1         & 1.2     \\
$\alpha=0.25$           & 2.74              & 0.50             & 3.3       & 18.7         & 1.2     \\
$\alpha=0.30$        & 2.82              & 0.48             & 2.9       & 17.4        & 1.0     \\
$\alpha=0.35$                  & 2.69              & 0.51             & 3.2       & 18.3         & 1.1     \\ 
$\alpha=0.40$                  & 2.76              & 0.49             & 3.5       & 19.2         & 1.2     \\\bottomrule
\end{tabular}
\label{tab:3}
%}
%\end{adjustbox}
\end{table}

\noindent \textbf{Exploration Intensity Controlling.}
Our method employs an intensity control factor $\alpha$ to weight the gradients of preferred and dispreferred samples in training. 
As shown in Tab. \ref{tab:3}, we examine the impact of different values of factor $\alpha$ on our method on LLaVA-v1.5-7B.
The results indicate that our method remains relatively robust to the control factor within an appropriate range, 
with minor performance differences observed among closely related values, and achieving relatively optimal performance at $\alpha=0.30$. 
Using the same selection strategy, we set $\alpha=0.25$ for LLaVA-v1.5-13B and $\alpha=0.15$ for Qwen2.5-VL-3B.
Overall, as the capability of the backbone model improves, the required exploration intensity control factor for effective hallucination suppression tends to decrease, which aligns well with intuitive expectations.

\begin{table}[b]
\centering
\caption{Ablation study on the visual sensitivity threshold $\tau$.}
\renewcommand{\arraystretch}{1.02}
\setlength{\tabcolsep}{1pt}
\small
%\begin{adjustbox}{max width=\textwidth}
%\resizebox{\textwidth}{!}{
\begin{tabular}{lccccc}
\toprule
\multirow{2}{*}{\textbf{Method}} & \multicolumn{2}{c}{\textbf{MMHal-Bench}} & \multicolumn{3}{c}{\textbf{AMBER}} \\ \cmidrule(r){2-3} \cmidrule(lr){4-6} 
                        & Score$\uparrow$         & HalRate$\downarrow$        & CHAIR$\downarrow$   & HalRate$\downarrow$   & Cog$\downarrow$   \\ \midrule
\emph{Sensitivity threshold}            &                &               &         &           &       \\
$\tau=0.3$                     & 2.63              & 0.53             & 3.3       & 18.8         & 1.3     \\
$\tau=0.4$           & 2.82              & 0.48             & 2.9       & 17.4        & 1.0     \\
$\tau=0.5$        & 2.74              & 0.50             & 3.2       & 18.2        & 1.1     \\ \bottomrule
\end{tabular}
\label{tab:4}
%}
%\end{adjustbox}
\end{table}

\begin{figure*}[ht]
    \begin{center}
    \includegraphics[width=17.4cm]{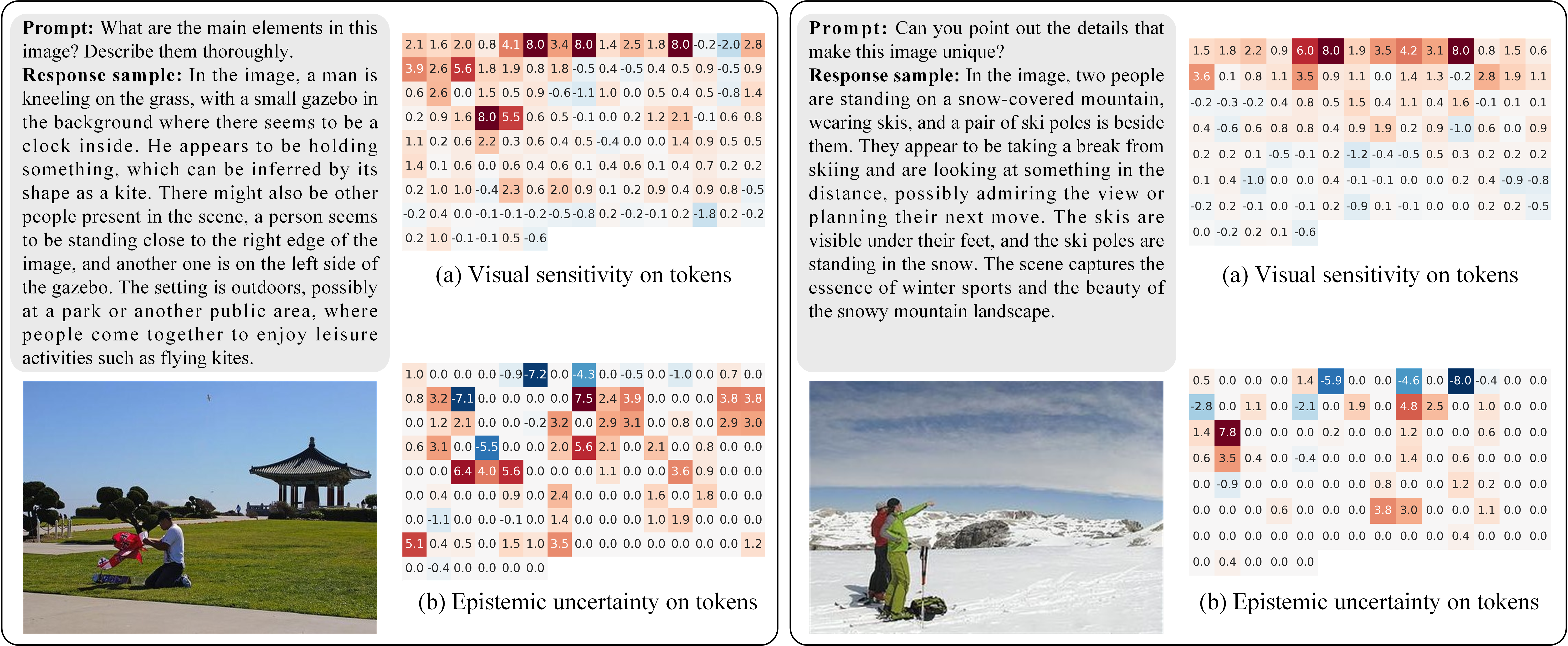}
    \caption{Visualization of the token-wise visual sensitivity and epistemic uncertainty in generating the corresponding responses. Positive values are indicated by red blocks and negative values by blue blocks, with deeper colors indicating larger absolute magnitudes at the corresponding positions. White denotes positions with a value of zero. The examples are ground-truth preferred samples from RLHF-V.} 
    \label{fig3}         
    \end{center}
\end{figure*}

\noindent \textbf{Visual Sensitivity Threshold.}
The visual sensitivity threshold $\tau$ specifies the proportion of tokens that are selected by the exploration mechanism for targeted optimization.
In essence, $\tau$ controls the ratio of tokens whose optimization pressure is adaptively adjusted in training.
For simplicity, a uniform threshold is adopted for both preferred and dispreferred samples in experiments.
As presented in Tab.~\ref{tab:4}, the ablation analysis indicates that a threshold value around $0.4$ yields the strongest performance.
Notably, prior credit self-estimation methods generally adjust the optimization of all tokens, 
whereas under UE-DPO, 
superior results are achieved by reassigning learning pressure to fewer than half.
This observation highlights the efficiency of UE-DPO's selective credit-assignment mechanism.

\subsection{Visualization}

To further investigate the intrinsic distinctions between the proposed epistemic uncertainty $\mathrm{u}$ (Sec. \ref{uncertainty}) and the conventional visual sensitivity, we visualize both signals through heatmaps for comparison, as shown in Fig. \ref{fig3}.
The visual sensitivity is quantified by $\Delta$ (Sec. \ref{sec:Exploration}),
which captures the variation in token logits induced by image blurring.

There are two primary observations that are consistent with our underlying motivation.
First, tokens exhibiting high visual sensitivity may correspond to low epistemic uncertainty, indicating that the associated visual information has been sufficiently internalized by the model.
Conversely, tokens with low visual sensitivity can still exhibit high epistemic uncertainty, suggesting that the related semantics remain insufficiently understood and should be assigned greater learning pressure for deeper exploratory refinement.
Secondly, the epistemic-uncertainty signal demonstrates greater sparsity, providing more targeted and effective guidance for directional exploration in learning than conventional visual sensitivity.
Notably, in the example on the left, although our method flags the relevant text portion of “other people” as relatively low sensitivity and high uncertainty, 
the shown responses are the ground-truth preferred samples and there are indeed other people in the distant background of the image.
Since such small objects elude model's perception capability and are therefore treated as its cognitive deficiencies, our method flags the relevant portion as low sensitivity and high uncertainty, and then triggers greater exploration to improve understanding.
This behavior is consistent with our motivation, although it may be constrained by model's perception bottleneck on tiny objects.

Taken together, 
the introduced epistemic uncertainty constitutes a valuable and theoretically grounded complement to visual sensitivity.
The integration of these two in UE-DPO enables more effective cognitive correction.

\section{Conclusion}
In this work, we introduce Uncertainty-aware Exploratory Direct Preference Optimization (UE-DPO), a novel method for mitigating hallucinations in multimodal large language models.
The proposed method guides the model to recognize its cognitive limitations and conduct self-refining exploration under token-level epistemic uncertainty.
By modeling uncertainty-aware exploration intensity, UE-DPO imposes greater learning pressure on visually deficient tokens in preferred samples while mitigating the over-penalization of beneficial visual knowledge in dispreferred ones.
Theoretically, we demonstrate that integrating the exploration intensity is equivalent to reformulate a generalized advantage function.
Extensive experiments on hallucination benchmarks, together with theoretical analysis, validate the effectiveness and robustness of our method.

\section*{Acknowledgments}

This research is supported by Artificial Intelligence-National Science and Technology Major Project 2023ZD0121200, Science Fund for Creative Research Groups under Grant 62121002,
and National Natural Science Foundation of China under Grant 62336001.

{
    \small
    \bibliographystyle{ieeenat_fullname}
    \bibliography{main}

\begin{thebibliography}{51}
\providecommand{\natexlab}[1]{#1}
\providecommand{\url}[1]{\texttt{#1}}
\expandafter\ifx\csname urlstyle\endcsname\relax
  \providecommand{\doi}[1]{doi: #1}\else
  \providecommand{\doi}{doi: \begingroup \urlstyle{rm}\Url}\fi

\bibitem[Bai et~al.(2025{\natexlab{a}})Bai, Cai, Chen, Chen, Chen, Cheng, Deng,
  Ding, Gao, Ge, et~al.]{bai2025qwen2}
Shuai Bai, Yuxuan Cai, Ruizhe Chen, Keqin Chen, Xionghui Chen, Zesen Cheng,
  Lianghao Deng, Wei Ding, Chang Gao, Chunjiang Ge, et~al.
\newblock Qwen2.5-vl technical report.
\newblock \emph{arXiv preprint arXiv:2502.13923}, 2025{\natexlab{a}}.

\bibitem[Bai et~al.(2025{\natexlab{b}})Bai, Cai, Chen, Chen, Chen, Cheng, Deng,
  Ding, Gao, Ge, et~al.]{bai2025qwen3}
Shuai Bai, Yuxuan Cai, Ruizhe Chen, Keqin Chen, Xionghui Chen, Zesen Cheng,
  Lianghao Deng, Wei Ding, Chang Gao, Chunjiang Ge, et~al.
\newblock Qwen3-vl technical report.
\newblock \emph{arXiv preprint arXiv:2511.21631}, 2025{\natexlab{b}}.

\bibitem[Che et~al.(2025)Che, Liu, Jia, Qin, Tang, and
  Pavlovic]{che2025hallucinatory}
Liwei Che, Tony~Qingze Liu, Jing Jia, Weiyi Qin, Ruixiang Tang, and Vladimir
  Pavlovic.
\newblock Hallucinatory image tokens: A training-free eazy approach to
  detecting and mitigating object hallucinations in lvlms.
\newblock In \emph{Proceedings of the IEEE/CVF International Conference on
  Computer Vision}, pages 21635--21644, 2025.

\bibitem[Chen et~al.(2025)Chen, Lyu, Gao, Song, and Shen]{chen2025attention}
Beitao Chen, Xinyu Lyu, Lianli Gao, Jingkuan Song, and Heng~Tao Shen.
\newblock Attention hijackers: Detect and disentangle attention hijacking in
  lvlms for hallucination mitigation.
\newblock \emph{arXiv preprint arXiv:2503.08216}, 2025.

\bibitem[Comanici et~al.(2025)Comanici, Bieber, Schaekermann, Pasupat,
  Sachdeva, Dhillon, Blistein, Ram, Zhang, Rosen, et~al.]{comanici2025gemini}
Gheorghe Comanici, Eric Bieber, Mike Schaekermann, Ice Pasupat, Noveen
  Sachdeva, Inderjit Dhillon, Marcel Blistein, Ori Ram, Dan Zhang, Evan Rosen,
  et~al.
\newblock Gemini 2.5: Pushing the frontier with advanced reasoning,
  multimodality, long context, and next generation agentic capabilities.
\newblock \emph{arXiv preprint arXiv:2507.06261}, 2025.

\bibitem[Dai et~al.(2023)Dai, Li, Li, Tiong, Zhao, Wang, Li, Fung, and
  Hoi]{dai2023instructblip}
Wenliang Dai, Junnan Li, Dongxu Li, Anthony Tiong, Junqi Zhao, Weisheng Wang,
  Boyang Li, Pascale~N Fung, and Steven Hoi.
\newblock Instructblip: Towards general-purpose vision-language models with
  instruction tuning.
\newblock \emph{Advances in neural information processing systems},
  36:\penalty0 49250--49267, 2023.

\bibitem[Gong et~al.(2024)Gong, Ming, Wang, and Wei]{gong2024damro}
Xuan Gong, Tianshi Ming, Xinpeng Wang, and Zhihua Wei.
\newblock Damro: Dive into the attention mechanism of lvlm to reduce object
  hallucination.
\newblock In \emph{Proceedings of the 2024 Conference on Empirical Methods in
  Natural Language Processing}, pages 7696--7712, 2024.

\bibitem[Gu et~al.(2024)Gu, Wang, Cao, Bu, Song, He, Li, and
  Zheng]{gu2024token}
Jihao Gu, Yingyao Wang, Meng Cao, Pi Bu, Jun Song, Yancheng He, Shilong Li, and
  Bo Zheng.
\newblock Token preference optimization with self-calibrated visual-anchored
  rewards for hallucination mitigation.
\newblock \emph{arXiv preprint arXiv:2412.14487}, 2024.

\bibitem[Hejna et~al.(2024)Hejna, Rafailov, Sikchi, Finn, Niekum, Knox, and
  Sadigh]{hejnacontrastive}
Joey Hejna, Rafael Rafailov, Harshit Sikchi, Chelsea Finn, Scott Niekum,
  W~Bradley Knox, and Dorsa Sadigh.
\newblock Contrastive preference learning: Learning from human feedback without
  reinforcement learning.
\newblock In \emph{The Twelfth International Conference on Learning
  Representations}, 2024.

\bibitem[Ho et~al.(2020)Ho, Jain, and Abbeel]{ho2020denoising}
Jonathan Ho, Ajay Jain, and Pieter Abbeel.
\newblock Denoising diffusion probabilistic models.
\newblock \emph{Advances in neural information processing systems},
  33:\penalty0 6840--6851, 2020.

\bibitem[Hu et~al.(2022)Hu, Shen, Wallis, Allen-Zhu, Li, Wang, Wang, and
  Chen]{hu2022lora}
Edward~J Hu, Yelong Shen, Phillip Wallis, Zeyuan Allen-Zhu, Yuanzhi Li, Shean
  Wang, Lu Wang, and Weizhu Chen.
\newblock Lo{RA}: Low-rank adaptation of large language models.
\newblock In \emph{International Conference on Learning Representations}, 2022.

\bibitem[Huang et~al.(2024)Huang, Dong, Zhang, Wang, He, Wang, Lin, Zhang, and
  Yu]{huang2024opera}
Qidong Huang, Xiaoyi Dong, Pan Zhang, Bin Wang, Conghui He, Jiaqi Wang, Dahua
  Lin, Weiming Zhang, and Nenghai Yu.
\newblock Opera: Alleviating hallucination in multi-modal large language models
  via over-trust penalty and retrospection-allocation.
\newblock In \emph{Proceedings of the IEEE/CVF Conference on Computer Vision
  and Pattern Recognition}, pages 13418--13427, 2024.

\bibitem[Hurst et~al.(2024)Hurst, Lerer, Goucher, Perelman, Ramesh, Clark,
  Ostrow, Welihinda, Hayes, Radford, et~al.]{hurst2024gpt}
Aaron Hurst, Adam Lerer, Adam~P Goucher, Adam Perelman, Aditya Ramesh, Aidan
  Clark, AJ Ostrow, Akila Welihinda, Alan Hayes, Alec Radford, et~al.
\newblock Gpt-4o system card.
\newblock \emph{arXiv preprint arXiv:2410.21276}, 2024.

\bibitem[Knox et~al.(2024)Knox, Hatgis-Kessell, Adalgeirsson, Booth, Dragan,
  Stone, and Niekum]{knox2024learning}
W~Bradley Knox, Stephane Hatgis-Kessell, Sigurdur~Orn Adalgeirsson, Serena
  Booth, Anca Dragan, Peter Stone, and Scott Niekum.
\newblock Learning optimal advantage from preferences and mistaking it for
  reward.
\newblock In \emph{Proceedings of the AAAI Conference on Artificial
  Intelligence}, pages 10066--10073, 2024.

\bibitem[Kuhn et~al.(2023)Kuhn, Gal, and Farquhar]{kuhn2023semantic}
Lorenz Kuhn, Yarin Gal, and Sebastian Farquhar.
\newblock Semantic uncertainty: Linguistic invariances for uncertainty
  estimation in natural language generation.
\newblock \emph{arXiv preprint arXiv:2302.09664}, 2023.

\bibitem[Lee et~al.(2024)Lee, Park, Jo, and Seo]{lee2024volcano}
Seongyun Lee, Sue Park, Yongrae Jo, and Minjoon Seo.
\newblock Volcano: Mitigating multimodal hallucination through self-feedback
  guided revision.
\newblock In \emph{Proceedings of the 2024 Conference of the North American
  Chapter of the Association for Computational Linguistics: Human Language
  Technologies (Volume 1: Long Papers)}, pages 391--404, 2024.

\bibitem[Leng et~al.(2024)Leng, Zhang, Chen, Li, Lu, Miao, and
  Bing]{leng2024mitigating}
Sicong Leng, Hang Zhang, Guanzheng Chen, Xin Li, Shijian Lu, Chunyan Miao, and
  Lidong Bing.
\newblock Mitigating object hallucinations in large vision-language models
  through visual contrastive decoding.
\newblock In \emph{Proceedings of the IEEE/CVF Conference on Computer Vision
  and Pattern Recognition}, pages 13872--13882, 2024.

\bibitem[Li et~al.(2025{\natexlab{a}})Li, Zhang, Jie, Ma, and
  Li]{li2025mitigating}
Jiaming Li, Jiacheng Zhang, Zequn Jie, Lin Ma, and Guanbin Li.
\newblock Mitigating hallucination for large vision language model by
  inter-modality correlation calibration decoding.
\newblock \emph{arXiv preprint arXiv:2501.01926}, 2025{\natexlab{a}}.

\bibitem[Li et~al.(2025{\natexlab{b}})Li, Shi, Gao, Liu, Wang, Chen, Liu, Zhao,
  Wang, and Metaxas]{li2025hidden}
Zhuowei Li, Haizhou Shi, Yunhe Gao, Di Liu, Zhenting Wang, Yuxiao Chen, Ting
  Liu, Long Zhao, Hao Wang, and Dimitris~N Metaxas.
\newblock The hidden life of tokens: Reducing hallucination of large
  vision-language models via visual information steering.
\newblock In \emph{International Conference on Machine Learning}, pages
  35799--35819, 2025{\natexlab{b}}.

\bibitem[Liu et~al.(2024{\natexlab{a}})Liu, Li, Li, and Lee]{liu2024improved}
Haotian Liu, Chunyuan Li, Yuheng Li, and Yong~Jae Lee.
\newblock Improved baselines with visual instruction tuning.
\newblock In \emph{Proceedings of the IEEE/CVF conference on computer vision
  and pattern recognition}, pages 26296--26306, 2024{\natexlab{a}}.

\bibitem[Liu et~al.(2024{\natexlab{b}})Liu, Zang, Dong, Zhang, Cao, Duan, He,
  Xiong, Lin, and Wang]{liu2024mia}
Ziyu Liu, Yuhang Zang, Xiaoyi Dong, Pan Zhang, Yuhang Cao, Haodong Duan,
  Conghui He, Yuanjun Xiong, Dahua Lin, and Jiaqi Wang.
\newblock Mia-dpo: Multi-image augmented direct preference optimization for
  large vision-language models.
\newblock \emph{arXiv preprint arXiv:2410.17637}, 2024{\natexlab{b}}.

\bibitem[Ma et~al.(2025)Ma, Chen, Zhou, Wang, and Zhang]{ma2025estimating}
Huan Ma, Jingdong Chen, Joey~Tianyi Zhou, Guangyu Wang, and Changqing Zhang.
\newblock Estimating llm uncertainty with evidence.
\newblock \emph{arXiv preprint arXiv:2502.00290}, 2025.

\bibitem[Peng et~al.(2025)Peng, Yang, Jiang, and Tian]{peng2025mitigating}
Shangpin Peng, Senqiao Yang, Li Jiang, and Zhuotao Tian.
\newblock Mitigating object hallucinations via sentence-level early
  intervention.
\newblock In \emph{Proceedings of the IEEE/CVF International Conference on
  Computer Vision}, pages 635--646, 2025.

\bibitem[Pi et~al.(2024)Pi, Han, Xiong, Zhang, Liu, Pan, and
  Zhang]{pi2024strengthening}
Renjie Pi, Tianyang Han, Wei Xiong, Jipeng Zhang, Runtao Liu, Rui Pan, and Tong
  Zhang.
\newblock Strengthening multimodal large language model with bootstrapped
  preference optimization.
\newblock In \emph{European Conference on Computer Vision}, pages 382--398,
  2024.

\bibitem[Rafailov et~al.(2023)Rafailov, Sharma, Mitchell, Manning, Ermon, and
  Finn]{rafailov2023direct}
Rafael Rafailov, Archit Sharma, Eric Mitchell, Christopher~D Manning, Stefano
  Ermon, and Chelsea Finn.
\newblock Direct preference optimization: Your language model is secretly a
  reward model.
\newblock \emph{Advances in neural information processing systems},
  36:\penalty0 53728--53741, 2023.

\bibitem[Rafailov et~al.(2024)Rafailov, Hejna, Park, and Finn]{rafailov2024r}
Rafael Rafailov, Joey Hejna, Ryan Park, and Chelsea Finn.
\newblock From \(r\) to \( q^{*}\): Your language model is secretly a
  q-function.
\newblock \emph{arXiv preprint arXiv:2404.12358}, 2024.

\bibitem[Rohrbach et~al.(2018)Rohrbach, Hendricks, Burns, Darrell, and
  Saenko]{rohrbach2018object}
Anna Rohrbach, Lisa~Anne Hendricks, Kaylee Burns, Trevor Darrell, and Kate
  Saenko.
\newblock Object hallucination in image captioning.
\newblock In \emph{Proceedings of the 2018 Conference on Empirical Methods in
  Natural Language Processing}, pages 4035--4045, 2018.

\bibitem[Sarkar et~al.(2024)Sarkar, Ebrahimi, Etemad, Beirami, Ar{\i}k, and
  Pfister]{sarkar2024mitigating}
Pritam Sarkar, Sayna Ebrahimi, Ali Etemad, Ahmad Beirami, Sercan~{\"O} Ar{\i}k,
  and Tomas Pfister.
\newblock Mitigating object hallucination in mllms via data-augmented
  phrase-level alignment.
\newblock \emph{arXiv preprint arXiv:2405.18654}, 2024.

\bibitem[Sun et~al.(2024)Sun, Shen, Cao, Liu, Li, Shen, Gan, Gui, Wang, Yang,
  et~al.]{sun2024aligning}
Zhiqing Sun, Sheng Shen, Shengcao Cao, Haotian Liu, Chunyuan Li, Yikang Shen,
  Chuang Gan, Liangyan Gui, Yu-Xiong Wang, Yiming Yang, et~al.
\newblock Aligning large multimodal models with factually augmented rlhf.
\newblock In \emph{Findings of the Association for Computational Linguistics:
  ACL 2024}, pages 13088--13110, 2024.

\bibitem[Suo et~al.(2025)Suo, Zhang, Sun, Wu, Wang, and Zhang]{suo2025octopus}
Wei Suo, Lijun Zhang, Mengyang Sun, Lin~Yuanbo Wu, Peng Wang, and Yanning
  Zhang.
\newblock Octopus: Alleviating hallucination via dynamic contrastive decoding.
\newblock In \emph{Proceedings of the Computer Vision and Pattern Recognition
  Conference}, pages 29904--29914, 2025.

\bibitem[Wang et~al.(2024{\natexlab{a}})Wang, Chen, Zhang, Tian, Xu, Deng, and
  Chen]{wang2024mllm}
Chenxi Wang, Xiang Chen, Ningyu Zhang, Bozhong Tian, Haoming Xu, Shumin Deng,
  and Huajun Chen.
\newblock Mllm can see? dynamic correction decoding for hallucination
  mitigation.
\newblock \emph{arXiv preprint arXiv:2410.11779}, 2024{\natexlab{a}}.

\bibitem[Wang et~al.(2024{\natexlab{b}})Wang, Zhou, Huang, Xu, Zhang, Poon, and
  Chen]{wang2024mdpo}
Fei Wang, Wenxuan Zhou, James~Y Huang, Nan Xu, Sheng Zhang, Hoifung Poon, and
  Muhao Chen.
\newblock mdpo: Conditional preference optimization for multimodal large
  language models.
\newblock In \emph{Proceedings of the 2024 Conference on Empirical Methods in
  Natural Language Processing}, pages 8078--8088, 2024{\natexlab{b}}.

\bibitem[Wang et~al.(2023)Wang, Wang, Xu, Zhang, Gu, Jia, Yan, Zhang, and
  Sang]{wang2023llm}
Junyang Wang, Yuhang Wang, Guohai Xu, Jing Zhang, Yukai Gu, Haitao Jia, Ming
  Yan, Ji Zhang, and Jitao Sang.
\newblock An llm-free multi-dimensional benchmark for mllms hallucination
  evaluation.
\newblock \emph{arXiv preprint arXiv:2311.07397}, 2023.

\bibitem[Wang et~al.(2024{\natexlab{c}})Wang, Pan, Ding, and
  Biemann]{wang2024mitigating}
Xintong Wang, Jingheng Pan, Liang Ding, and Chris Biemann.
\newblock Mitigating hallucinations in large vision-language models with
  instruction contrastive decoding.
\newblock In \emph{Findings of the Association for Computational Linguistics:
  ACL 2024}, pages 15840--15853, 2024{\natexlab{c}}.

\bibitem[Xie et~al.(2024)Xie, Li, Xu, and Kan]{xie2024v}
Yuxi Xie, Guanzhen Li, Xiao Xu, and Min-Yen Kan.
\newblock V-dpo: Mitigating hallucination in large vision language models via
  vision-guided direct preference optimization.
\newblock In \emph{Findings of the Association for Computational Linguistics:
  EMNLP 2024}, pages 13258--13273, 2024.

\bibitem[Yadkori et~al.(2024)Yadkori, Kuzborskij, Gy{\"o}rgy, and
  Szepesvari]{yadkori2024believe}
Yasin~A Yadkori, Ilja Kuzborskij, Andr{\'a}s Gy{\"o}rgy, and Csaba Szepesvari.
\newblock To believe or not to believe your llm: Iterative prompting for
  estimating epistemic uncertainty.
\newblock \emph{Advances in Neural Information Processing Systems},
  37:\penalty0 58077--58117, 2024.

\bibitem[Yang et~al.(2025)Yang, Luo, Han, Xu, and Li]{yang2025mitigating}
Zhihe Yang, Xufang Luo, Dongqi Han, Yunjian Xu, and Dongsheng Li.
\newblock Mitigating hallucinations in large vision-language models via dpo:
  On-policy data hold the key.
\newblock In \emph{Proceedings of the Computer Vision and Pattern Recognition
  Conference}, pages 10610--10620, 2025.

\bibitem[Yin et~al.(2025)Yin, Si, and Wang]{yin2025clearsight}
Hao Yin, Guangzong Si, and Zilei Wang.
\newblock Clearsight: Visual signal enhancement for object hallucination
  mitigation in multimodal large language models.
\newblock In \emph{Proceedings of the Computer Vision and Pattern Recognition
  Conference}, pages 14625--14634, 2025.

\bibitem[Yin et~al.(2024)Yin, Fu, Zhao, Xu, Wang, Sui, Shen, Li, Sun, and
  Chen]{yin2024woodpecker}
Shukang Yin, Chaoyou Fu, Sirui Zhao, Tong Xu, Hao Wang, Dianbo Sui, Yunhang
  Shen, Ke Li, Xing Sun, and Enhong Chen.
\newblock Woodpecker: Hallucination correction for multimodal large language
  models.
\newblock \emph{Science China Information Sciences}, 67\penalty0 (12):\penalty0
  220105, 2024.

\bibitem[Yu et~al.(2024)Yu, Yao, Zhang, He, Han, Cui, Hu, Liu, Zheng, Sun,
  et~al.]{yu2024rlhf}
Tianyu Yu, Yuan Yao, Haoye Zhang, Taiwen He, Yifeng Han, Ganqu Cui, Jinyi Hu,
  Zhiyuan Liu, Hai-Tao Zheng, Maosong Sun, et~al.
\newblock Rlhf-v: Towards trustworthy mllms via behavior alignment from
  fine-grained correctional human feedback.
\newblock In \emph{Proceedings of the IEEE/CVF Conference on Computer Vision
  and Pattern Recognition}, pages 13807--13816, 2024.

\bibitem[Yu et~al.(2025)Yu, Zhang, Li, Xu, Yao, Chen, Lu, Cui, Dang, He,
  et~al.]{yu2025rlaif}
Tianyu Yu, Haoye Zhang, Qiming Li, Qixin Xu, Yuan Yao, Da Chen, Xiaoman Lu,
  Ganqu Cui, Yunkai Dang, Taiwen He, et~al.
\newblock Rlaif-v: Open-source ai feedback leads to super gpt-4v
  trustworthiness.
\newblock In \emph{Proceedings of the Computer Vision and Pattern Recognition
  Conference}, pages 19985--19995, 2025.

\bibitem[Zhang et~al.(2025)Zhang, Wan, Kan, Ma, Stepputtis, Ramanan,
  Salakhutdinov, Morency, Sycara, and Xie]{zhang2025self}
Ce Zhang, Zifu Wan, Zhehan Kan, Martin~Q Ma, Simon Stepputtis, Deva Ramanan,
  Russ Salakhutdinov, Louis-Philippe Morency, Katia Sycara, and Yaqi Xie.
\newblock Self-correcting decoding with generative feedback for mitigating
  hallucinations in large vision-language models.
\newblock \emph{arXiv preprint arXiv:2502.06130}, 2025.

\bibitem[Zhang et~al.(2024)Zhang, Quan, Gu, Shen, Yuan, Yan, Cheng, Wu, and
  Ye]{zhang2024seeing}
Xiaofeng Zhang, Yihao Quan, Chaochen Gu, Chen Shen, Xiaosong Yuan, Shaotian
  Yan, Hao Cheng, Kaijie Wu, and Jieping Ye.
\newblock Seeing clearly by layer two: Enhancing attention heads to alleviate
  hallucination in lvlms.
\newblock \emph{arXiv preprint arXiv:2411.09968}, 2024.

\bibitem[Zhao et~al.(2025)Zhao, Si, Chen, Zhang, Sun, Chang, and
  Zhang]{zhao2025looking}
Haozhe Zhao, Shuzheng Si, Liang Chen, Yichi Zhang, Maosong Sun, Baobao Chang,
  and Minjia Zhang.
\newblock Looking beyond text: Reducing language bias in large vision-language
  models via multimodal dual-attention and soft-image guidance.
\newblock In \emph{Proceedings of the 2025 Conference on Empirical Methods in
  Natural Language Processing}, pages 19677--19701, 2025.

\bibitem[Zhao et~al.(2024)Zhao, Deng, Zhang, and Gu]{zhao2024mitigating}
Linxi Zhao, Yihe Deng, Weitong Zhang, and Quanquan Gu.
\newblock Mitigating object hallucination in large vision-language models via
  classifier-free guidance.
\newblock \emph{arXiv preprint arXiv:2402.08680}, 2024.

\bibitem[Zhao et~al.(2023)Zhao, Wang, Ouyang, Dong, Wang, and
  He]{zhao2023beyond}
Zhiyuan Zhao, Bin Wang, Linke Ouyang, Xiaoyi Dong, Jiaqi Wang, and Conghui He.
\newblock Beyond hallucinations: Enhancing lvlms through hallucination-aware
  direct preference optimization.
\newblock \emph{arXiv preprint arXiv:2311.16839}, 2023.

\bibitem[Zhou et~al.(2023)Zhou, Cui, Yoon, Zhang, Deng, Finn, Bansal, and
  Yao]{zhou2023analyzing}
Yiyang Zhou, Chenhang Cui, Jaehong Yoon, Linjun Zhang, Zhun Deng, Chelsea Finn,
  Mohit Bansal, and Huaxiu Yao.
\newblock Analyzing and mitigating object hallucination in large
  vision-language models.
\newblock \emph{arXiv preprint arXiv:2310.00754}, 2023.

\bibitem[Zhou et~al.(2024)Zhou, Cui, Rafailov, Finn, and Yao]{zhou2024aligning}
Yiyang Zhou, Chenhang Cui, Rafael Rafailov, Chelsea Finn, and Huaxiu Yao.
\newblock Aligning modalities in vision large language models via preference
  fine-tuning.
\newblock \emph{arXiv preprint arXiv:2402.11411}, 2024.

\bibitem[Zhu et~al.(2025{\natexlab{a}})Zhu, Ji, Chen, Xu, Ye, and
  Liu]{zhu2025ibd}
Lanyun Zhu, Deyi Ji, Tianrun Chen, Peng Xu, Jieping Ye, and Jun Liu.
\newblock Ibd: Alleviating hallucinations in large vision-language models via
  image-biased decoding.
\newblock In \emph{Proceedings of the Computer Vision and Pattern Recognition
  Conference}, pages 1624--1633, 2025{\natexlab{a}}.

\bibitem[Zhu et~al.(2025{\natexlab{b}})Zhu, Tao, Dong, and
  Xu]{zhu2025mitigating}
Younan Zhu, Linwei Tao, Minjing Dong, and Chang Xu.
\newblock Mitigating object hallucinations in large vision-language models via
  attention calibration.
\newblock \emph{arXiv preprint arXiv:2502.01969}, 2025{\natexlab{b}}.

\bibitem[Zhuang et~al.(2025)Zhuang, Zhu, Xie, Liang, and
  Zou]{zhuang2025vasparse}
Xianwei Zhuang, Zhihong Zhu, Yuxin Xie, Liming Liang, and Yuexian Zou.
\newblock Vasparse: Towards efficient visual hallucination mitigation via
  visual-aware token sparsification.
\newblock In \emph{Proceedings of the Computer Vision and Pattern Recognition
  Conference}, pages 4189--4199, 2025.

\end{thebibliography}
}

\clearpage
\setcounter{page}{1}
\maketitlesupplementary

\appendix

\section{The Implicit Advantage View of DPO}

The learning objective in the KL-regularized RL setting, such as DPO, is to find a policy $\pi_\theta$ that maximizes the expected sum of rewards while penalizing its deviation from a fixed reference policy $\pi_\mathrm{ref}$,
which can be formally expressed as the maximization of the objective:
{\small
\begin{equation}
J(\pi)=\mathbb{E}_{\tau \sim \pi}\left[\sum_{t=0}^T\left(r\left(s_t, a_t\right)-\beta D_{\mathrm{KL}}\left(\pi\left(\cdot \mid s_t\right) \| \pi_\mathrm{ref}\left(\cdot \mid s_t\right)\right)\right)\right],
\label{eq:klrlapp}
\end{equation}}%
where $\tau$ represents a trajectory,
$r(s_t, a_t)$ is the reward function,
and $\beta > 0$ is a temperature parameter that controls the strength of the KL-divergence penalty.

Connecting the trajectory-level optimization problem to single-step decision-making,
the KL-regularized Bellman relations between the optimal state-value $V^*$ and optimal action-value $Q^*$ functions are as follows:
\begin{equation}
V^*(s)=\max _{\pi(\cdot \mid s)}\left\{\mathbb{E}_{a \sim \pi}\left[Q^*(s, a)\right]-\beta D_{\mathrm{KL}}(\pi \| \pi_\mathrm{ref})\right\},
\label{bellman1}
\end{equation}
\begin{equation}
Q^*(s, a)=r(s, a)+\gamma \mathbb{E}_{s^{\prime}}\left[V^*\left(s^{\prime}\right)\right],
\label{bellman2}
\end{equation}
where $s^{\prime}$ denotes the subsequent state reached after state $s$ receives token $a$.
Eq. \ref{bellman1} and \ref{bellman2} defines the local objective at state $s$:
to find a policy $\pi(\cdot|s)$ that maximizes the expected Q$^*$-value while remaining close to the reference policy $\pi_{\text{ref}}$. 
The discount factor $\gamma$ is typically set to $1$ in large language model settings.
Moreover,
since the state transition in language modeling is deterministic,
Eq. \ref{bellman2} can be simplified as
$Q^*(s, a)=r(s, a)+ V^*\left(s^{\prime}\right)$.

From the perspective of dynamic programming, 
global optimality is guaranteed only if every state $s$ satisfies the local optimality condition.
Equivalently, violations at any single state imply sub-optimality at the global level. 
This principle formalizes the transition from the global optimization of $J(\pi)$ to the hierarchy of per-state optimizations.
To solve the per-state optimization:
\begin{equation}
\max _{\pi(\cdot \mid s)}\left\{\mathbb{E}_{a \sim \pi}\left[Q^*(s, a)\right]-\beta D_{K L}(\pi \| \pi_\mathrm{ref})\right\},
\end{equation}
subject to the constraint that $\pi(\cdot|s)$ is a valid probability distribution, \emph{i.e.}:
\begin{equation}
    \sum_a \pi(a|s) = 1.
\end{equation}
The method of Lagrange multipliers can be used to solve this constrained problem:
{\small
\begin{equation}
\begin{aligned}
\mathcal{L}(\pi,\eta)
&=
\left(
\sum_a \pi(a|s) Q^*(s,a)
- \beta \sum_a \pi(a|s) \log \frac{\pi(a|s)}{\pi_{\mathrm{ref}}(a|s)}
\right)
\\
&\quad
- \eta \left( \sum_a \pi(a|s) - 1 \right).
\end{aligned}
\end{equation}}
To find the optimal policy,
taking the partial derivative of $\mathcal{L}$ with respect to $\pi(a|s)$ and setting it to $0$:
\begin{equation}
Q^*(s, a)-\beta\left(\log \frac{\pi^*(a \vert s)}{\pi_{\mathrm{ref}}(a \vert s)}+1\right) - \eta =0.
\end{equation}
Solving for $\pi^*(a|s)$ yields:
\begin{equation}
\pi^*(a \vert s)=\pi_{\mathrm{ref}}(a \vert s) \exp \left(\frac{Q^*(s, a)}{\beta}\right) \exp \left(\frac{-\eta(s)-\beta}{\beta}\right).
\end{equation}
The term $\exp((-\eta(s) - \beta)/\beta)$ is constant with respect to any token (action) $a$. 
We can thus absorb it into a state-dependent normalization term, $1/Z(s)$, 
which ensures that the policy sums to $1$ over all tokens.
This gives the closed-form expression for the optimal policy $\pi^*(a|s)$:
\begin{equation}
\pi^*(a \vert s)=\frac{1}{Z(s)} \pi_{\mathrm{ref}}(a \vert s) \exp \left(\frac{Q^*(s, a)}{\beta}\right),
\label{eq:piop}
\end{equation}
where the partition function $Z(s)$ is defined by the normalization constraint: 
\begin{equation}
Z(s) = \sum_{a'} \pi_{\text{ref}}(a'|s) \exp(Q^*(s, a')/\beta).
\end{equation}
From the derivation of the optimal policy, we can rearrange Eq. \ref{eq:piop} to express $\log(\pi^*(a|s)/\pi_{\text{ref}}(a|s))$:
\begin{equation}
\log \frac{\pi^*(a \vert s)}{\pi_{\mathrm{ref}}(a \vert s)}=\frac{Q^*(s, a)}{\beta}-\log Z(s).
\label{eq:piratio}
\end{equation}
By substituting Eq. \ref{eq:piratio} back into the Bellman equation for $V^*(s)$, the expression can be simplified to:
\begin{equation}
V^*(s)=\beta \log Z(s).
\end{equation}

\begin{algorithm*}[t]
\caption{Uncertainty-aware Exploratory Direct Preference Optimization (UE-DPO)}
\label{alg:ue-dpo}
\begin{algorithmic}[1]
\Require Dataset $\mathcal{D}=\{(v, x, y_w, y_l)\}$, policy model $\pi_\theta$, reference model $\pi_{\text{ref}}$
\Require Hyperparameters: noise level $\xi$, quantile $\tau$, intensity scale $\alpha$, learning rate $\eta$
\While{not converged}
    \State Sample a batch $\{(v, x, y_w, y_l)\}$ from $\mathcal{D}$
    \State \textbf{// Uncertainty Awareness (Sec.~4.1)}
    \State Generate blurred image $v'$ via diffusion noise:
        $v'(k) \leftarrow \sqrt{\bar{\xi}_k} \cdot v + \sqrt{1-\bar{\xi}_k} \cdot \epsilon$
    \State Compute token logit variation $\Delta(a_t, s_t) \leftarrow \mathrm{logit}_\theta(a_t \vert v,x,y_{<t}) - \mathrm{logit}_\theta(a_t \vert v^\prime,x,y_{<t})$
    \State Compute token epistemic uncertainty $\mathrm{u}(a_t, s_t) \leftarrow \mathrm{logit}_\theta(\hat{a}_t(v^\prime) \vert v,x,y_{<t}) - \mathrm{logit}_\theta(a_t \vert v,x,y_{<t})$
    \State \textbf{// The Control of Exploration Intensity (Sec.~4.2)}
    \State Preferred branch ($y_w$):
    \State Identify visually insensitive mask: $I_w \leftarrow 1\{\Delta(a_t,s_t) \le q_\tau(\Delta)\}$
    \State Compute exploration intensity: $\lambda_w(a_t, s_t) \leftarrow 1 + \alpha \cdot I_w \cdot \sigma(\text{Normalize}(\mathrm{u}))$
    \State Dispreferred branch ($y_l$):
    \State Identify visually sensitive mask: $I_l \leftarrow 1\{\Delta(a_t,s_t) \ge q_{1-\tau}(\Delta)\}$
    \State Compute exploration intensity: $\lambda_l(a_t, s_t) \leftarrow 1 - \alpha \cdot I_l \cdot \sigma(\text{Normalize}(\mathrm{u}))$
    \State \textbf{// Training Objective (Sec.~4.3)}
    \State Compute UE-DPO loss:
        \[
        \mathcal{L}_{\text{UE-DPO}} \leftarrow -\mathbb{E} \left[ 
            \log \sigma \left(
                \beta \sum_t \log \frac{\pi_\theta(a^w_t|s^w_t)^{\lambda_w}}{\pi_{\text{ref}}(a^w_t|s^w_t)}
                - \beta \sum_t \log \frac{\pi_\theta(a^l_t|s^l_t)^{\lambda_l}}{\pi_{\text{ref}}(a^l_t|s^l_t)}
            \right)
        \right]
        \]
    \State Update $\theta \leftarrow \theta - \eta \nabla_\theta \mathcal{L}_{\text{UE-DPO}}$
\EndWhile
\end{algorithmic}
\end{algorithm*}

Finally, 
we can derive an explicit expression for the optimal advantage function, defined as $A^*(s, a) \triangleq Q^*(s, a) - V^*(s)$. 
By substituting $V^*(s) = \beta \log Z(s)$ into the right-hand side of Eq. \ref{eq:piratio}, we arrive at the key relationship:
\begin{equation}
\beta \log \left(\frac{\pi^*(a \vert s)}{\pi_{\mathrm{ref}}(a \vert s)}\right)=Q^*(s, a)-V^*(s) \triangleq A^*(s, a)
\end{equation}
The quantity $A^*(s, a)$ measures the improvement in expected performance of token (action) $a$ over the reference baseline at state $s$, that is, how much better token $a$ performs compared to the baseline performance under reference policy $\pi_{\mathrm{ref}}$.
This relationship is central to the DPO framework, 
where the advantage is parameterized as an implicit immediate reward, 
and policy improvement is conducted through the supervised alignment on preference pairs.

DPO parameterizes this advantage as an implicit immediate reward and performs preference learning by fitting pairwise preference data $\left(x, y_w, y_l\right) \sim \mathcal{D}$:
\begin{equation}
\begin{aligned}
& L_{\mathrm{DPO}}\left(\pi_\theta, \pi_{\mathrm{ref}}\right)=-\mathbb{E}_{\left(x, y_w, y_l\right) \sim \mathcal{D}} \\
& \log \sigma\left(\beta \sum_{t=0}^{T_w} \log \frac{\pi_\theta\left(a_t^w \vert s_t\right)}{\pi_{\mathrm{ref}}\left(a_t^w \vert s_t\right)}-\beta \sum_{t=0}^{T_l} \log \frac{\pi_\theta\left(a_t^l \vert s_t\right)}{\pi_{\mathrm{ref}}\left(a_t^l \vert s_t\right)}\right).
\end{aligned}
\end{equation}

\section{Algorithmic Outline}

We summarize our method in Alg. \ref{alg:ue-dpo}.

\section{The Influence of Exploration Intensity on the Optimal Policy}

The integration of the uncertainty-aware exploration can be interpreted as introducing a token-wise entropy regularization factor $\lambda(s, a)$ into the standard reverse-KL regularized RL objective in Eq. \ref{eq:klrlapp}:
(Note that, for notational simplicity, we henceforth denote $\lambda(s, a)$ and $\lambda(s, a^\prime)$ as $\lambda$ and $\lambda^\prime$, respectively, in the subsequent equations.)
\begin{equation}
\begin{aligned}
& \mathbb{E}_{a \sim \pi} \left[\log \pi(a \vert s)-\log \pi_{\mathrm{ref}}(a \vert s)\right] \\
& \rightarrow \mathbb{E}_{a \sim \pi} \left[\lambda \log \pi(a \vert s)-\log \pi_{\mathrm{ref}}(a \vert s)\right].
\end{aligned}
\end{equation}
This factor dynamically modulates the KL regularization strength in accordance with the model's epistemic uncertainty across tokens, 
leading to the following reformulated optimal value function:
\begin{equation}
\max _{\pi(\cdot \vert s)}\{\mathbb{E}_{a \sim \pi}\left[Q^*(s, a) - \beta (\lambda \log \pi(a \vert s) - \log \pi_{\mathrm{ref}}(a \vert s))\right]\}.
\label{eq:ueobjapp}
\end{equation}
Subsequently, the Lagrangian function is constructed as follows:
{\small
\begin{equation}
\begin{aligned}
L(\pi, \eta)
&= \sum_a  \pi(a \vert s) [Q^*(s, a) 
   - \beta (\lambda \log \pi(a \vert s) - \log  \pi_\mathrm{ref}(a \vert s))]\\
& + \eta \left(\sum_a \pi(a \vert s) - 1\right).
\end{aligned}
\end{equation}}%
Taking the partial derivative with respect to $\pi(a \vert s)$ and equating it to zero yields:
\begin{equation}
\frac{\partial L}{\partial \pi(a \vert s)}  = Q^*- \beta \lambda \left(1 + \log \pi \right)+\beta \log \pi_\mathrm{ref} +\eta = 0
\end{equation}
The optimal policy can be derived as:
\begin{equation}
\pi^{*}(a \vert s)=\pi_{\mathrm{ref}}(a \vert s)^{1 / \lambda} \exp \left( \frac{Q^{*}(s, a) + \eta(s)}{\beta \lambda}\right) / Z(s),
\label{eq:revisedobjapp}
\end{equation}
where $Z(s)$ is partition function, and $\eta$ is Lagrange multiplier.
$\eta(s)$ indicates that the multiplier $\eta$ depends only on the state $s$ and is independent of token $a$.

The partial derivative of $\log \pi^{*}$ with respect to $\lambda(s,a)$ is:
\begin{equation}
\frac{\partial}{\partial \lambda} \log \pi^{*} \propto  -\frac{\log \left(\pi_{\mathrm{ref}}(a \vert s)\right)} {\lambda^2} - \frac{\left(Q^*(s,a)+\eta(s)\right)}{\left(\beta \lambda^2\right)},
\end{equation}
where \textbf{the first term $-\log \left(\pi_{\mathrm{ref}}(a \vert s)\right) / \lambda^2$} is always positive, since the prior probability $\pi_\mathrm{ref}$ takes value in the interval between 0 and 1. 
This term, for tokens $a$ with a small $\pi_\mathrm{ref}$, exerts a relatively strong positive driving force, 
thereby endowing the factor $\pi_{\mathrm{ref}}(a \vert s)^{1 / \lambda}$ in Eq. \ref{eq:revisedobjapp} with a prior-corrective exploratory effect.
\textbf{The second term $ -\left(Q^*(s,a)+\eta(s)\right) / (\beta \lambda^2)$} is negative for the valuable tokens of interest, that is, those with large $Q^*(s,a)$.
This term exerts a pulling force that attempts to reduce their probability through reducing the $\exp(\cdot)$ factor in  Eq. \ref{eq:revisedobjapp}, 
such that the $\exp(\cdot)$ factor contributes to a smoothing effect on $\pi^{*}$.
This smoothing effect questions the authority of high-$Q^*$ tokens and induces a conservative, broad exploratory behavior.
\textbf{The overall exploration mechanism} of $\lambda$ is precisely the outcome of the dynamic interplay between these two effects. 
The ultimate sign of $\frac{\partial}{\partial \lambda} \log \pi^{*}$ depends on the relative magnitudes of $-\log \left(\pi_{\mathrm{ref}}(a \vert s)\right) / \lambda^2$ and $ \left(Q^*(s,a)+\eta(s)\right) / (\beta \lambda^2)$.

For the ``forgotten correct tokens'' of interest in our UE-DPO framework,
namely, those that are correct but have been omitted by the reference policy, with small $\pi_\mathrm{ref}$ yet large $Q^*$,
the term $-\log \left(\pi_{\mathrm{ref}}(a \vert s)\right) / \lambda^2$ dominates $ \left(Q^*(s,a)+\eta(s)\right) / (\beta \lambda^2)$, that is,
$\pi_{\mathrm{ref}}(a \vert s) < \exp \left(-(Q^*(s,a) + \eta(s))/\beta\right)$.
In this case, since the derivative is positive, increasing $\lambda$ raises the probability of these tokens in the optimized policy $\pi^*$, 
embodying an error-correcting mode of exploration.

\section{Generalized Exploratory Advantage}

It follows that the optimal policy $\pi^{*}(a \vert s)$ ( in Eq. \ref{eq:revisedobjapp}) satisfies:
\begin{equation}
\beta \log \frac{\pi^*(a \vert s)^{\lambda}}{\pi_{\mathrm{ref}}(a \vert s)} = Q^{*}(s,a) - \beta \lambda(s,a) + \eta(s),
\label{eq:conditionapp}
\end{equation}
Recall that the optimal value function $V^*(s)$ in Eq. \ref{eq:ueobjapp}:
\begin{equation}
V^* = \mathbb{E}_{a \sim \pi^*}\left[Q^*(s,a) - \beta (\lambda \log \pi^*(a \vert s) - \log \pi_{\mathrm{ref}}(a \vert s))\right].
\label{eq:optvapp}
\end{equation}
Substituting Eq. \ref{eq:conditionapp} into Eq. \ref{eq:optvapp}, we have:
\begin{equation}
V^*(s) = \beta \mathbb{E}_{a^\prime \sim \pi^*}\left[ \lambda(s, a^\prime)\right] - \eta(s).
\label{eq:optvsapp}
\end{equation}
Then, substituting Eq. \ref{eq:optvsapp} into Eq. \ref{eq:conditionapp}, we obtain:
\begin{equation}
\begin{aligned}
&\beta \log \frac{\pi^*(a \vert s)^{\lambda}}{\pi_{\mathrm{ref}}(a \vert s)} \\
& = Q^{*}(s,a)-V^{*}(s)-\beta (\lambda -\mathbb{E}_{a^{\prime} \sim \pi^*}\left[\lambda^\prime\right]) \\
& \triangleq A_\mathrm{e}^*(s,a),
\end{aligned}
\label{eq:conditionf}
\end{equation}
which we refer to as the generalized exploratory advantage $A_\mathrm{e}^*$.
This generalized advantage $A_\mathrm{e}^*$ consists of two components: the \textbf{standard advantage function} $Q^{*}(s,a)-V^{*}(s)$ and a novel \textbf{exploration cost advantage} $-\beta (\lambda -\mathbb{E}_{a^{\prime} \sim \pi^*}\left[\lambda^\prime\right])$.
If $\lambda(a,s)$ is higher than the average cost $\mathbb{E}_{a^{\prime} \sim \pi^*}\left[\lambda^\prime\right]$, 
\emph{i.e.}, $\lambda - \mathbb{E}_{a^{\prime} \sim \pi^*}\left[\lambda^\prime\right] > 0$,
which implies that the token $a$ receives a stronger exploration cost penalty, thereby reducing its total advantage $ A_\mathrm{e}^*(s,a)$.
As the intensity $\lambda(s,a)$ diminishes in preference learning, the exploratory advantage $A_\mathrm{e}^*$ increases.

By substituting this generalized exploratory advantage $A_\mathrm{e}^*$ into DPO framework,
our proposed UE-DPO method can fit the pair-wise preference ordering while effectively mitigating the under-cognition of visual information through the $\lambda$-weighted gradients for active exploration, as discussed in the main paper.

\end{document}